\PassOptionsToPackage{table}{xcolor}
\PassOptionsToPackage{pagebackref,breaklinks=true,colorlinks,citecolor=blue,urlcolor=blue,linkcolor=blue,bookmarks=false}{hyperref}
\PassOptionsToPackage{noabbrev,nameinlink,capitalize}{cleveref}

\def\arxivroot{}
\IfFileExists{arxiv/april_aigc.cls}{%
  \def\arxivroot{arxiv/}%
  \documentclass[onecolumn, numbers]{arxiv/april_aigc}%
}{%
  \documentclass[onecolumn, numbers]{april_aigc}%
}

\ifdefined\pdfobjcompresslevel
\fi

\graphicspath{{\arxivroot}}
\usepackage[utf8]{inputenc} 
\usepackage{url}            
\usepackage{amsfonts}       
\usepackage{amssymb}        
\usepackage{amsmath}        
\usepackage{nicefrac}       
\usepackage{microtype}      
\usepackage{xspace}         
\usepackage{fix-cm}
\usepackage[T1]{fontenc}

\usepackage{booktabs}       
\usepackage{wrapfig}        
\usepackage{needspace}      
\usepackage{multicol}       
\usepackage{multirow}       
\usepackage{makecell}       
\usepackage{tabularx}       
\usepackage{adjustbox}      
\usepackage{longtable}      
\usepackage{colortbl}       
\usepackage{pifont}         
\usepackage{enumitem}
\usepackage[normalem]{ulem}

\usepackage{xcolor}
\definecolor{linkcolor}{named}{aprilblue}
\definecolor{urlcolor}{RGB}{255,105,180}
\definecolor{citecolor}{RGB}{66,168,235}
\definecolor{lightgray}{rgb}{0.8, 0.8, 0.8}
\definecolor{darkgreen}{rgb}{0.00, 0.81, 0.78}

\definecolor{gray_tab}{RGB}{220, 220, 220}
\definecolor{blue_tab}{RGB}{227, 240, 251}
\definecolor{oran_tab}{RGB}{252, 242, 237}
\definecolor{whit_tab}{RGB}{255, 255, 255}
\definecolor{green_code}{RGB}{55, 126, 34}
\definecolor{codeblue}{RGB}{40,75,160}
\definecolor{codekw}{RGB}{150,45,45}

\usepackage{algorithm}
\usepackage{algorithmic}
\usepackage{listings}
\usepackage{etoolbox}

\makeatletter
\AfterEndEnvironment{algorithm}{\let\@algcomment\relax}
\AtEndEnvironment{algorithm}{\kern2pt\hrule\relax\vskip3pt\@algcomment}
\let\@algcomment\relax
\newcommand\algcomment[1]{\def\@algcomment{\footnotesize#1}}
\renewcommand\fs@ruled{\def\@fs@cfont{\bfseries}\let\@fs@capt\floatc@ruled
  \def\@fs@pre{\hrule height.8pt depth0pt \kern2pt}%
  \def\@fs@post{}%
  \def\@fs@mid{\kern2pt\hrule\kern2pt}%
  \let\@fs@iftopcapt\iftrue}
\makeatother

\newcommand{\cmark}{\ding{52}\xspace}%
\newcommand{\xmarkg}{\textcolor{lightgray}{\ding{56}}\xspace}%
\usepackage[pagebackref,breaklinks=true,colorlinks,citecolor=blue,urlcolor=blue,linkcolor=blue,bookmarks=false]{hyperref}
\AtEndPreamble{
    \usepackage[capitalize]{cleveref}
    \crefname{section}{Sec.}{Secs.}
    \Crefname{section}{Section}{Sections}
    \crefname{table}{Tab.}{Tabs.}
    \Crefname{table}{Table}{Tables}
    \crefname{equation}{Eq.}{Eqs.}
    \Crefname{equation}{Equation}{Equations}
    \crefname{figure}{Fig.}{Figs.}
    \Crefname{figure}{Figure}{Figures}
    \crefname{lstlisting}{Listing}{Listings}
    \Crefname{lstlisting}{Listing}{Listings}
}
\hypersetup{colorlinks=true,linkcolor=linkcolor,urlcolor=urlcolor,citecolor=citecolor}

\usepackage{caption}
\DeclareCaptionFormat{custom}{{\color{aprilblue}\sffamily\textbf{#1 #2}} #3}
\titleformat*{\section}{\color{aprilblue}\Large\sffamily\bfseries}
\titleformat*{\subsection}{\color{aprilblue}\large\sffamily\bfseries}
\titleformat*{\subsubsection}{\color{aprilblue}\normalsize\sffamily\bfseries}

\usepackage{fancyhdr}
\newif\ifshowlogo
\showlogotrue   
\newcommand{\insertlogo}{%
  \ifshowlogo
    \IfFileExists{\arxivroot assets/april_logo1.png}%
    {\includegraphics[height=0.68cm]{\arxivroot 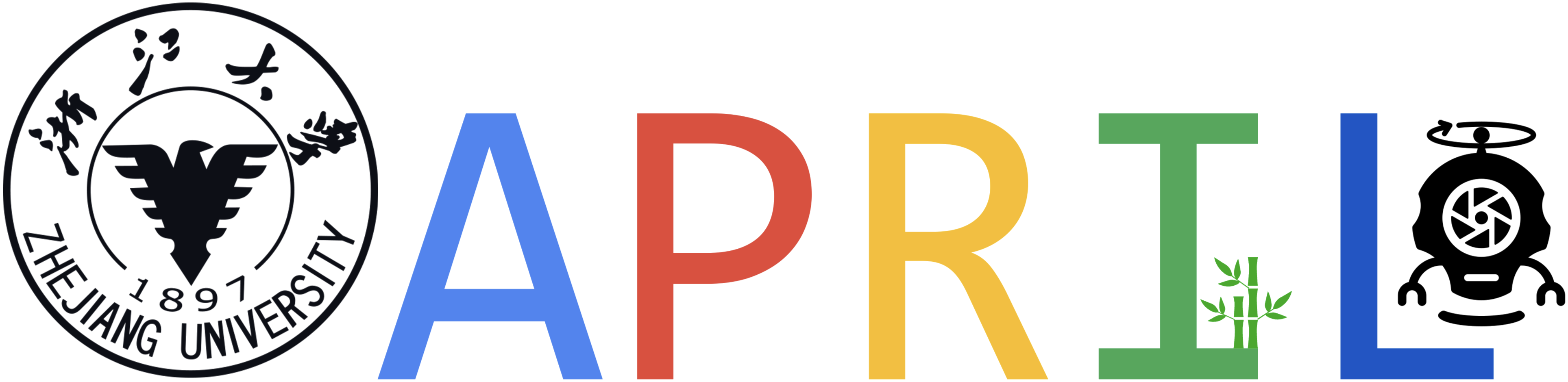}}
    {}%
  \fi
}
\newif\ifshowtoc
\showtocfalse

\def\method{DAS}
\def\dataset{DAS-2M}
\def\benchmark{DAS-Bench}
\def\evalname{DAS-Eval}

\renewcommand{\title}[1]{\def\titlelist{{\fontsize{20pt}{28pt}\selectfont\sffamily\bfseries #1}}}
\title{Deep Academic Survey:\\ Stateful Agentic Closed-Loop Paradigm for Academic Survey Automation}

\author[1,\star]{Zhikai Xu}
\author[1,\star]{Zhucun Xue}
\author[2]{Teng Hu}
\author[1]{Yabiao Wang}
\author[1]{Yong Liu}
\author[1,\dagger]{Jiangning Zhang}

\affiliation[1]{Zhejiang University}
\affiliation[2]{Shanghai Jiao Tong University}

\contribution[\star]{Equal contribution}
\contribution[\dagger]{Corresponding author}

\abstract{Academic surveys play a central role in organizing rapidly expanding scholarly literature, yet their construction requires extensive paper analysis, coherent knowledge organization, fine-grained citation support, and reliable manuscript assembly. Existing Deep Research and automated survey generation systems address parts of this process, but typically do not coordinate paper understanding, literature organization, evidence-grounded drafting, and manuscript validation through a shared, revisable state. We introduce \method{}, a stateful agentic framework for generating publication-oriented academic surveys. Its key idea is to separate reusable paper analysis from topic-specific manuscript construction. \method{} builds on \dataset{}, a dynamically updated metadata lake containing survey-oriented representations of approximately two million papers. Its agents maintain explicit literature, organization, writing, and finalization states through candidate-grounded taxonomy planning, reverse paper-to-section routing, and hierarchical claim and citation planning. Semantic review reactivates only the affected writing states for repair and reevaluation, forming a scoped closed loop with deterministic validation. We further introduce \benchmark{}, a 30-topic benchmark, together with \evalname{}, which assesses scholarly citation quality, taxonomic synthesis, hierarchical discourse, and manuscript assembly reliability through 16 criteria. Among systems evaluated on all 30 topics, \method{} achieves the highest average in all four dimensions, with an overall score of 4.34 compared with 4.03 for the strongest competitor, and the same ordering is preserved on the matched 21-topic CS subset. Blinded expert evaluation further prefers \method{} to Naive RAG on 27 of 30 topics and to AutoSurvey on 19 of 21 shared CS topics.
}

\coverdate{\today}
\covercorrespondence{\email{186368@zju.edu.cn}}
\coversourcecode{https://github.com/ZhikaiXu24/DAS}
\coverdataset{https://huggingface.co/datasets/ZhikaiXu24/DAS-2M}
\coverproject{https://zhikaixu24.github.io/projects/DAS/}

\begin{document}

\maketitle
\thispagestyle{plain}

\ifshowtoc
    \clearpage
    \setcounter{tocdepth}{2} 
    
    \tableofcontents
    \vspace{1cm} 


    \clearpage
\fi

\section{Introduction}

The rapid growth of scholarly literature has made high-quality academic surveys essential entry points for understanding unfamiliar fields, organizing their intellectual structure, and tracking recent advances \cite{bornmann2015growth}. Yet producing such surveys manually requires broad literature coverage, continuous updating, and substantial expert effort, making the process increasingly difficult to sustain as the literature expands \cite{snyder2019literature}.

\begin{wrapfigure}{R}{0.54\textwidth}
    \centering
    \includegraphics[width=\linewidth]{\arxivroot 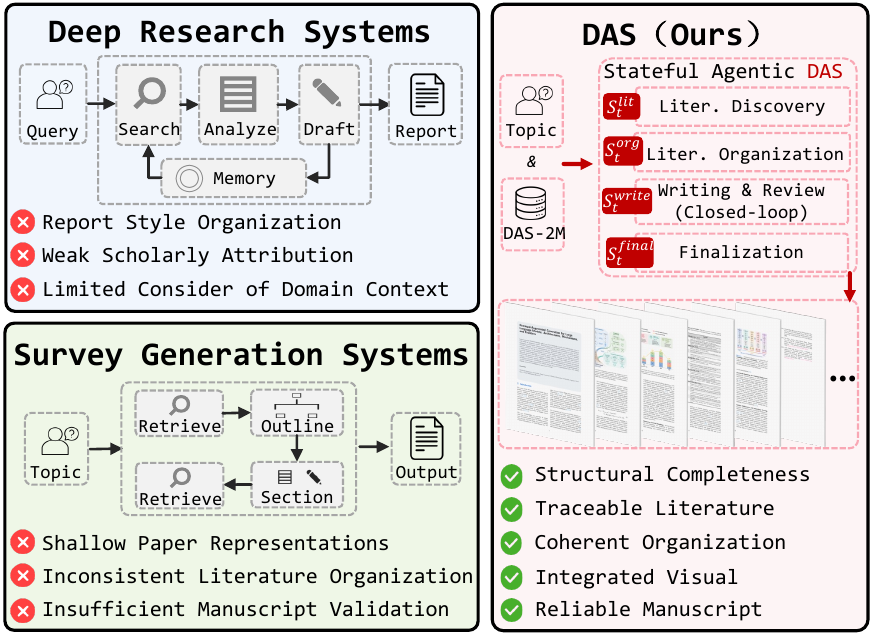}
    \caption{\textbf{Overview of existing systems and their limitations.} Deep Research and automated survey generation systems typically center on retrieval and drafting, but do not jointly support the literature representation, organization, and validation required for publication-oriented academic surveys.}
    \label{fig:teaser}
\end{wrapfigure}

We define publication-oriented survey generation as constructing an academic survey with structural completeness, traceable literature support, coherent organization across taxonomy, section, and paragraph levels, integrated visual elements, and consistency among final manuscript artifacts \cite{fu2025reproducible}. As illustrated in \cref{fig:teaser}, we contrast \method{} with two categories of existing systems: Deep Research systems and automated survey generation systems. \textbf{\textit{1) Deep Research Systems.}} Systems such as OpenAI Deep Research and Gemini Deep Research support broad retrieval and long-form report generation with citations, but are designed primarily for general research assistance rather than academic survey construction \cite{openai2025deepresearch,google2024deepresearch}. \textbf{\textit{i) Report-Style Organization.}} Their outputs may lack a coherent taxonomy, sustained academic exposition, and integrated equations, tables, and figures. \textbf{\textit{ii) Weak Scholarly Attribution.}} References may mix academic and web sources, follow nonstandard formats, or align imprecisely with supported claims \cite{gao2023alce,asai2026openscholar}. \textbf{\textit{iii) Limited Consideration of Domain Context.}} Domain constraints, experimental settings, and comparable evaluation dimensions may be insufficiently analyzed. \textbf{\textit{2) Automated Survey Generation Systems.}} These systems provide stronger scholarly focus and more structured outputs \cite{wang2024autosurvey,yan2025surveyforge,go2026lira}, but still face three challenges. \textbf{\textit{i) Shallow or Recomputed Paper Representations.}} Titles and abstracts omit technical and empirical details, while repeated document-level extraction introduces substantial processing cost \cite{dasigi2021qasper}. \textbf{\textit{ii) Inconsistent Literature Organization.}} Topic-only outline generation, independent section-level retrieval, and direct drafting can lead to inconsistent coverage, section boundaries, paper assignments, and claim-level citation support. \textbf{\textit{iii) Insufficient Manuscript Validation.}} Publication-oriented surveys require both semantic review for coherence and deterministic validation for structural integrity and compilability, yet existing systems rarely support both.

To address these challenges, we propose \textbf{Deep Academic Survey (\method{})}, a stateful agentic framework for constructing publication-oriented survey manuscripts. \textbf{First,} \method{} constructs \dataset{}, a persistent and dynamically updated literature metadata lake that provides reusable, survey-oriented representations of approximately 2 million papers. \textbf{Second,} \method{} organizes the candidate literature into a taxonomy with section-aligned paper assignments, and then progressively translates these assignments into claim-level evidence support through hierarchical planning and drafting. \textbf{Finally,} \method{} employs a semantic review-and-repair loop to preserve logical coherence, together with deterministic validation to ensure the structural integrity and reliable assembly of the final manuscript. Together, these designs formulate survey generation as a structured, stateful, and closed-loop manuscript construction process. The complete methodology is presented in \cref{sec:agentic_das}.
Our contributions are summarized as follows:
\begin{itemize}
\item We propose \textbf{\method{}}, the first stateful agentic framework for generating publication-oriented academic surveys. We also construct \dataset{}, a literature metadata lake that provides fine-grained and survey-oriented paper representations for topic-specific agentic manuscript construction.

\item We design a closed-loop manuscript construction methodology comprising candidate-grounded taxonomy planning and reverse paper-to-section routing, hierarchical paragraph and claim-level citation planning, and a scoped semantic review-and-repair loop coupled with deterministic validation. Together, these mechanisms maintain cross-level consistency from candidate literature organization and section-level paper assignments to claim-level evidence support and final manuscript assembly.

\item We construct \benchmark{}, the first benchmark designed to evaluate publication-oriented academic survey generation. Comprehensive comparisons show that \method{} achieves the strongest overall performance among the compared systems.
\end{itemize}
\FloatBarrier

\section{Related Work}
\subsection{Deep Research for Scholarly Inquiry}
Retrieval-augmented and agentic systems increasingly support scholarly inquiry \cite{yao2023react}. RAG grounds generation in external evidence \cite{lewis2020rag}, while STORM extends retrieval to multi-perspective research and long-form writing \cite{shao2024storm}. Scientific agents such as PaperQA2 and OpenScholar further support iterative literature search and citation-grounded scientific question answering at scale \cite{skarlinski2024paperqa2,asai2026openscholar}. Commercial Deep Research agents similarly conduct multistep web investigation and produce cited reports \cite{google2024deepresearch,openai2025deepresearch}. Claude Science integrates literature analysis, scientific tools, computing resources, and artifact generation within an auditable research workbench \cite{anthropic2026claudescience}. However, these systems primarily target open-ended inquiry and cited reporting. \method{} instead formulates academic survey generation as stateful manuscript construction over explicit literature, organization, writing, and finalization states.

\subsection{Automated Survey Generation}
Systems for automated survey generation address this more specialized target \cite{hu2014related,wang2019paperrobot,lu2020multixscience,shi2023unified}. AutoSurvey and SurveyForge retrieve papers, construct outlines, draft survey content, and refine the resulting text, with SurveyForge incorporating outline heuristics derived from human practice and memory-guided scholarly navigation \cite{wang2024autosurvey,yan2025surveyforge,bao2025surveygen}. SurveyX organizes references through AttributeTrees, while STRUCTSURVEY introduces structured agentic retrieval based on entities, relations, and topical taxonomies \cite{liang2025surveyx,pedinotti2026structsurvey}. InteractiveSurvey exposes intermediate artifacts for user revision, and IterSurvey iteratively updates retrieval results and outlines \cite{wen2025interactivesurvey,zhang2025itersurvey}. LiRA coordinates specialized agents for outlining, subsection writing, editing, and review, while ARISE introduces multi-agent review with explicit evaluation rubrics \cite{go2026lira,wang2025arise,zhang2025minigraph}. DeepSurvey further combines full-paper analysis, section-level paper assignment, evidence-constrained citation generation, and multi-granularity refinement \cite{yang2026deepsurvey}.
\begin{table*}[!t]
\centering
\setlength{\tabcolsep}{2.5pt}
\renewcommand{\arraystretch}{1.12}
\small
\def\capsystem#1{\makebox[2.25cm][l]{#1}}
\def\capcorpus#1{\makebox[2.15cm][c]{#1}}
\def\capcol#1{\makebox[1.20cm][c]{#1}}
\resizebox{\textwidth}{!}{%
\begin{tabular}{@{}lc*{9}{c}@{}}
\toprule
\raisebox{-0.8ex}[0pt][0pt]{\capsystem{\textbf{System}}}
& \multicolumn{2}{c}{\textbf{Scholarly Substrate}}
& \multicolumn{3}{c}{\textbf{Structure and Grounding}}
& \multicolumn{2}{c}{\textbf{Revision and Validation}}
& \multicolumn{3}{c}{\textbf{Manuscript Artifacts}} \\
\cmidrule(lr){2-3}
\cmidrule(lr){4-6}
\cmidrule(lr){7-8}
\cmidrule(lr){9-11}
& \capcorpus{\textbf{Corpus}}
& \capcol{\textbf{PRep}}
& \capcol{\textbf{GTax}}
& \capcol{\textbf{LRoute}}
& \capcol{\textbf{DPlan}}
& \capcol{\textbf{RLoop}}
& \capcol{\textbf{DCheck}}
& \capcol{\textbf{VInt}}
& \capcol{\textbf{CAVis}}
& \capcol{\textbf{PDF}} \\
\midrule
\capsystem{AutoSurvey}
& \capcorpus{530K}
& \xmarkg & \xmarkg & \xmarkg & \xmarkg
& \xmarkg & \xmarkg
& \xmarkg & \xmarkg & \xmarkg \\
\capsystem{SurveyForge}
& \capcorpus{600K+20K}
& \xmarkg & \xmarkg & \xmarkg & \xmarkg
& \xmarkg & \xmarkg
& \xmarkg & \xmarkg & \xmarkg \\
\capsystem{SurveyX}
& \capcorpus{2.63M$^{*}$+online}
& \xmarkg & \cmark & \xmarkg & \xmarkg
& \xmarkg & \xmarkg
& \cmark & \xmarkg & \cmark \\
\capsystem{InteractiveSurvey}
& \capcorpus{online+uploads}
& \xmarkg & \cmark & \xmarkg & \xmarkg
& \xmarkg & \xmarkg
& \cmark & \xmarkg & \cmark \\
\capsystem{LiRA}
& \capcorpus{provided refs.}
& \xmarkg & \xmarkg & \xmarkg & \xmarkg
& \cmark & \xmarkg
& \xmarkg & \xmarkg & \xmarkg \\
\capsystem{DeepSurvey}
& \capcorpus{online}
& \xmarkg & \cmark & \cmark & \xmarkg
& \cmark & \xmarkg
& \xmarkg & \xmarkg & \xmarkg \\
\midrule
\capsystem{\textbf{\method{}}}
& \capcorpus{\textbf{2M}}
& \cmark & \cmark & \cmark & \cmark
& \cmark & \cmark
& \cmark & \cmark & \cmark \\
\bottomrule
\end{tabular}%
}
\caption{\textbf{Capabilities of representative automated survey generation systems.} $\checkmark$ and $\times$ denote documented and absent or undocumented capabilities, respectively. $^{*}$SurveyX reports an unreleased 2.63M-paper corpus.}
\label{tab:system_capabilities}
\end{table*}

Despite these advances, existing systems generally cover only part of the end-to-end process required for publication-oriented survey generation. \method{} addresses this gap through a unified stateful agentic framework that integrates candidate-grounded literature organization, hierarchical manuscript construction, and a scoped semantic review-and-repair loop coupled with deterministic finalization.

\begin{figure*}[!t]
    \centering
    \includegraphics[width=\textwidth]{\arxivroot 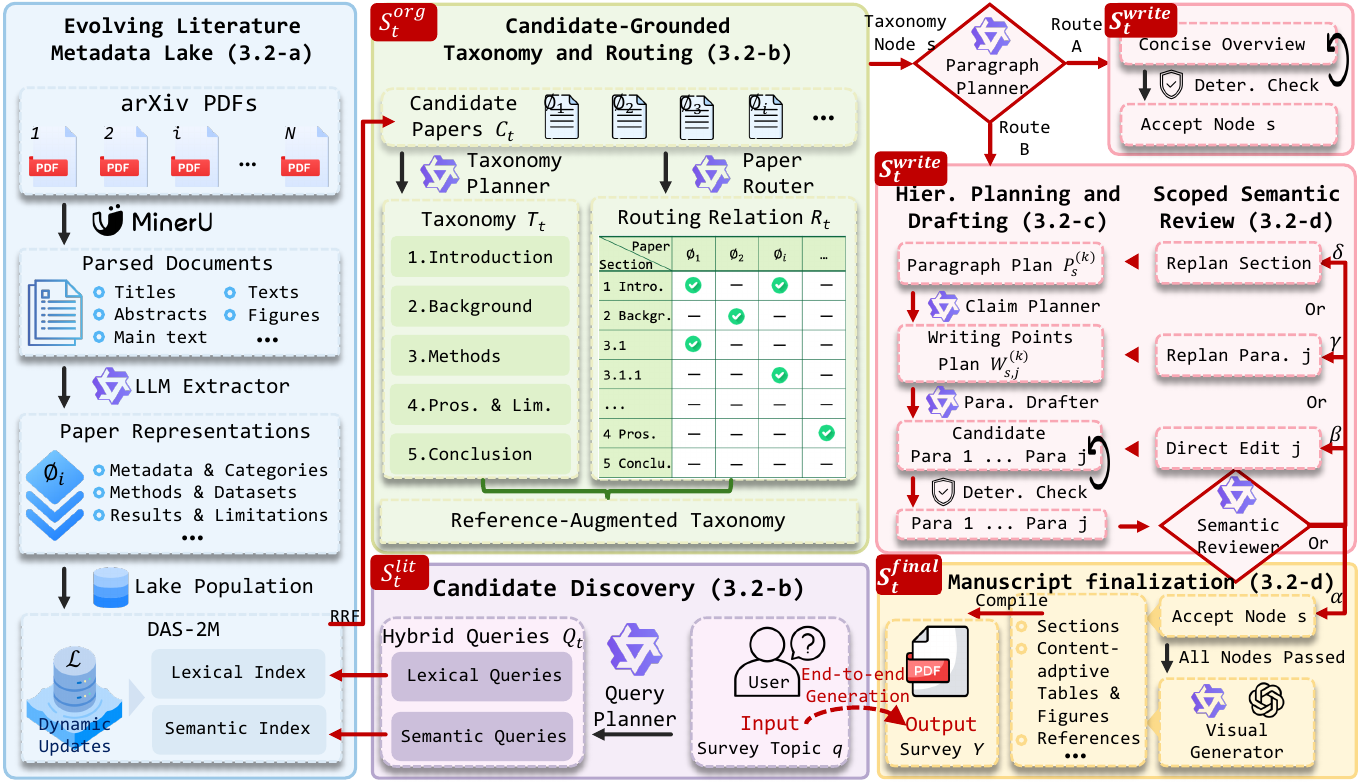}
    \caption{\textbf{Overview of the \method{} framework.} \method{} builds on the dynamically updated \dataset{} metadata lake and coordinates literature discovery, literature organization, hierarchical drafting, scoped review and repair, and artifact finalization through an explicit manuscript state. Review feedback reactivates the affected writing state before accepted content enters final assembly.}
    \label{fig:das_overview}
\end{figure*}

\section{Method}
\label{sec:method}
In this section, we first formulate the task of publication-oriented survey generation and define the shared manuscript state. We then present the methodology of \method{} and finally introduce \benchmark{}. \Cref{tab:system_capabilities} compares \method{} with representative automated survey generation systems in terms of their documented capabilities.
\subsection{Task Formulation}
\label{sec:task_formulation}
Given a survey topic $q$, a generation configuration $\Gamma$, and a dynamically updated literature metadata lake $\mathcal{L}=\{\phi_i\}_{i=1}^{N}$, where $\phi_i$ denotes the survey-oriented structured representation of paper $i$, \method{} performs stateful agentic construction to produce a publication-oriented academic survey manuscript $Y$.

Throughout this process, \method{} maintains a shared manuscript state $\mathcal{S}_t$ at construction step $t$:
\begin{equation}
\mathcal{S}_t =
\left(
\mathcal{S}^{\mathrm{lit}}_t,
\mathcal{S}^{\mathrm{org}}_t,
\mathcal{S}^{\mathrm{write}}_t,
\mathcal{S}^{\mathrm{final}}_t
\right),
\label{eq:manuscript_state}
\end{equation}
where $\mathcal{S}^{\mathrm{lit}}_t$ stores the query plan $\mathcal{Q}_t$ and candidate papers  $\mathcal{C}_t$, while $\mathcal{S}^{\mathrm{org}}_t$ stores the taxonomy $\mathcal{T}_t$ and paper-to-section routing relation $\mathcal{R}_t$. $\mathcal{S}^{\mathrm{write}}_t$ stores the paragraph plans, writing-point plans, optional source evidence, drafts, and review status associated with the taxonomy nodes, while $\mathcal{S}^{\mathrm{final}}_t$ stores the figures, tables, bibliographic records, and manuscript source files produced during finalization. These states follow an explicit dependency: the candidate literature informs taxonomy construction, the taxonomy and routing relation constrain drafting, and accepted drafts enter manuscript assembly. When review identifies a defect, \method{} reactivates only the affected writing states at the corresponding construction stage and re-executes the dependent agentic steps, thereby forming a scoped review-and-repair loop.

\subsection{Methodology: Agentic \method{}}
\label{sec:agentic_das}

We introduce \method{}, the first stateful agentic system designed for publication-oriented survey generation. As illustrated in \cref{fig:das_overview}, \method{} separates paper understanding from topic-specific manuscript construction. Its methodology comprises four components: a dynamically updated literature metadata lake, candidate-grounded taxonomy planning and paper routing, hierarchical planning and drafting, and a scoped semantic review-and-repair loop followed by manuscript finalization. Together, these components realize a stateful agentic closed-loop paradigm for academic survey automation.
\FloatBarrier

\subsubsection{Evolving Literature Metadata Lake}
\label{sec:metadata_lake}

\leavevmode\par
\noindent\textbf{Survey-oriented paper representation.}
A title-abstract pair provides only a coarse paper representation because technical mechanisms, implementation details, empirical findings, and limitations are often absent or heavily condensed \cite{dasigi2021qasper}. This limited view constrains fine-grained paper understanding and can reduce survey writing to shallow summaries of abstract-level information. In a retrieve-then-extract design applied separately to each topic, the same paper may undergo document-level LLM extraction repeatedly across different survey topics, resulting in redundant computation and increased generation latency. \method{} therefore precomputes a survey-oriented representation $\phi_i$ for each paper $i$ and stores it in $\mathcal{L}$. Each representation $\phi_i$ is organized into eight high-level field groups: bibliographic metadata, topical categorization, technical configuration, resource availability, methodological details, dataset usage, research rationale and findings, and empirical evaluation and limitations. The resulting structured representations are shared across the subsequent stages of survey construction.

\noindent\textbf{Literature metadata lake construction and maintenance.}
We construct $\mathcal{L}$ from approximately 2 million arXiv papers submitted between January 2020 and June 2026. The construction and maintenance process consists of four stages. \textbf{\textit{1) Full PDF parsing.}} We use MinerU to parse the complete content and document structure of each PDF, including titles, abstracts, the main text and other document elements \cite{wang2024mineru}. \textbf{\textit{2) Survey-oriented metadata extraction.}} An LLM-based extractor converts the parsed content into the survey-oriented paper representation defined above, capturing the information required by subsequent agentic taxonomy planning, paper routing, claim planning, drafting, and review. \textbf{\textit{3) Retrieval indexing.}} We construct lexical and semantic indexes over the extracted metadata to support hybrid candidate discovery for topic-specific survey construction. \textbf{\textit{4) Dynamic lake maintenance.}} The lake is dynamically updated through the same processing pipeline as new papers become available, allowing \method{} to incorporate recent literature and preserve the timeliness of generated surveys. The complete metadata schema, corpus statistics, extraction configuration, and quality-control procedure are provided in \cref{app:das2m}.

\subsubsection{Candidate-Grounded Taxonomy and Routing}
\label{sec:taxonomy_routing}
\leavevmode\par
\noindent\textbf{Hybrid candidate discovery.}
Given a survey topic $q$, the Query Planner agent expands the topic into complementary lexical and semantic queries and records them in $\mathcal{Q}_t$. A deterministic hybrid retriever executes BM25 \cite{robertson2009bm25} and dense search \cite{karpukhin2020dpr} over the offline indexes of $\mathcal{L}$ and merges the ranked results using weighted reciprocal rank fusion \cite{cormack2009rrf}. The highest-ranked papers, together with their retrieval provenance, form the candidate set $\mathcal{C}_t$. This transition updates $\mathcal{S}^{\mathrm{lit}}_t$ and provides the shared candidate literature for taxonomy planning.

\noindent\textbf{Candidate-grounded taxonomy planning.}
Given $\mathcal{C}_t$, the Taxonomy Planner agent reads a joint view of the structured paper representations and organizes the research directions covered by the candidate literature into a rooted taxonomy $\mathcal{T}_t$. Each node defines a semantic scope, a structural role, and a writing objective; nodes are designated as analytical, reflective, or navigational according to their function. A deterministic structural check verifies the hierarchy and normalizes section identifiers before $\mathcal{T}_t$ is stored in $\mathcal{S}^{\mathrm{org}}_t$. These node definitions guide both paper routing and subsequent drafting.

\noindent\textbf{Reverse paper-to-section routing.}
Because $\mathcal{C}_t$ is optimized for recall, it may contain weakly related papers and does not specify which sections each paper can support. Rather than retrieving papers independently for each section, the Paper Router agent evaluates each candidate representation against the eligible nodes of $\mathcal{T}_t$ using its technical focus, empirical findings, and research logic. The router produces a sparse multi-label assignment: a paper may be assigned to multiple nodes when it supports distinct section objectives, or to none when no sufficiently aligned node is found. The accepted assignments define
$\mathcal{R}_t\subseteq\mathcal{C}_t\times\operatorname{Nodes}(\mathcal{T}_t)$. Together, $\mathcal{T}_t$ and $\mathcal{R}_t$ complete $\mathcal{S}^{\mathrm{org}}_t$ and establish section-level citation scopes for subsequent drafting.

\subsubsection{Hierarchical Planning and Drafting}
\label{sec:hierarchical_drafting}

\leavevmode\par
\noindent\textbf{Adaptive paragraph planning.}
For each taxonomy node $s$, the Paragraph Planner agent reads its semantic scope, structural role, papers assigned through $\mathcal{R}_t$, and compact views of their structured representations. Based on this information, the planner selects one of two writing routes: \textbf{\textit{1) Route A.}} Navigational nodes and nodes whose assigned literature cannot support a multi-paragraph technical discussion follow Route A and produce a concise overview. \textbf{\textit{2) Route B.}} The remaining nodes enter hierarchical planning, drafting, and the scoped review-and-repair loop. For each Route B node, the planner constructs an initial ordered paragraph plan $\mathcal{P}_s^{(0)}$, whose entries specify paragraph themes, argumentative roles, target lengths, and paragraph-level paper assignments drawn from the node's routed citation scope. This plan separates paragraph responsibilities and defines the paper set available to each paragraph for subsequent claim and citation planning. For Route B nodes, $k$ indexes local construction and review iterations, with $k=0$ denoting the initial planning and drafting pass.

\noindent\textbf{Claim and citation planning.}
For each planned paragraph $j$, the Claim Planner agent reads its paragraph objective and the full structured representations of the papers assigned by $\mathcal{P}_s^{(k)}$. It constructs $\mathcal{W}_{s,j}^{(k)}$ as an ordered sequence of writing points, each specifying an intended claim, its supporting citation group, and the technical details required for drafting \cite{gao2023alce}. Papers that jointly support the same claim may share a citation group, while the same paper may be reused across writing points only when it supports a distinct aspect in each case. When an essential detail is unavailable in the structured representation, the planner may issue a bounded, focused request to the corresponding source document \cite{gao2023rarr}. Any retrieved evidence is attached to the relevant writing point in $\mathcal{W}_{s,j}^{(k)}$.

\noindent\textbf{Paragraph realization and validation.}
For each Route-B paragraph, the Drafter agent reads the section objective, paragraph theme, claim-level writing plan $\mathcal{W}_{s,j}^{(k)}$, attached source evidence, and preceding validated paragraphs. The Drafter generates a paragraph candidate, which is immediately passed to a deterministic validator before it can be committed to $\mathcal{S}^{\mathrm{write}}_t$. The deterministic checker verifies citation identifiers, unresolved placeholders, paragraph boundaries, section formatting, and LaTeX constraints. Violations that admit rule-based correction are resolved directly; otherwise, explicit revision instructions reactivate the Drafter for another generation step. Once validated, Route-A nodes produce a single validated overview. For Route-B nodes, validated paragraphs are assembled according to the current paragraph plan and submitted to the semantic review loop.

\subsubsection{Scoped Review and Manuscript Finalization}
\label{sec:scoped_review}

\leavevmode\par
\noindent\textbf{Scoped semantic review-and-repair loop.}
For each Route-B node $s$, the Reviewer operates on the subsection-specific portion of $\mathcal{S}^{\mathrm{write}}_t$ at local review iteration $k$, including the paragraph plan $\mathcal{P}_s^{(k)}$, the claim-level writing and citation plans $\{\mathcal{W}_{s,j}^{(k)}\}_{j=1}^{J_s^{(k)}}$, and the validated paragraph drafts $\{d_{s,j}^{(k)}\}_{j=1}^{J_s^{(k)}}$.

The Reviewer agent evaluates the assembled subsection $D_s^{(k)}$ for technical relevance, alignment with the taxonomy-defined objective, argumentative progression, redundancy across paragraphs, and citation support for central claims. According to the scope of the identified defect, the Reviewer selects one of four actions: accepting the current subsection, directly revising paragraph $j$, replanning paragraph $j$, or replanning the entire subsection. We denote these actions by $\alpha$, $\beta_j$, $\gamma_j$, and $\delta$, respectively.
The local writing state is updated as

\begin{equation}
\left(\mathcal{S}^{\mathrm{write}}_t\right)_{s}^{(k+1)}
=
\begin{cases}
\left(\mathcal{S}^{\mathrm{write}}_t\right)_{s}^{(k)},
& a_s^{(k)}=\alpha,\\[2pt]
F_{a_s^{(k)}}\!\left(\left(\mathcal{S}^{\mathrm{write}}_t\right)_{s}^{(k)}\right),
& a_s^{(k)}\neq\alpha.
\end{cases}
\label{eq:scoped_review_loop}
\end{equation}
where $F_{a_s^{(k)}}$ denotes the scoped repair operation applied to the subsection-specific writing state selected by the current review action. Action $\beta_j$ revises only the affected paragraph draft. Action $\gamma_j$ preserves the section-level paragraph plan while regenerating the corresponding claim-level writing and citation plan and paragraph draft. Action $\delta$ reconstructs the paragraph plan and all dependent writing states and may therefore change the number of planned paragraphs from $J_s^{(k)}$ to $J_s^{(k+1)}$.
Every regenerated paragraph must pass deterministic validation before being recommitted to the shared manuscript state. The subsection is then reassembled and evaluated again by the Reviewer agent. This repeated state update and re-evaluation forms a scoped agentic review-and-repair loop \cite{madaan2023selfrefine,shinn2023reflexion}, which continues until the subsection is accepted or the retry budget is exhausted. If the budget is exhausted, \method{} retains the latest complete version that has passed deterministic validation as the fallback writing state.

\noindent\textbf{Manuscript finalization.}
After drafting and review, specialized visual generation roles generate content-adaptive figures and tables from the accepted manuscript state. The finalizer then resolves bibliographic records, inserts and validates cross-references, and assembles the corresponding LaTeX, BibTeX, and asset files. These components update $\mathcal{S}^{\mathrm{final}}_t$, which is compiled into the final survey manuscript $Y$. Complete role prompts, structured output schemas, validation rules, and retry configurations are provided in \cref{app:das_implementation}.

\subsection{\benchmark{}}

\noindent\textbf{Benchmark construction.}
\benchmark{} contains 30 survey topics, comprising 21 core computer science topics and nine non-CS topics. All systems receive the same survey task. Closed-source systems use their native configurations, while reproducible systems retain their released workflows and use Qwen3.5-397B-FP8 as the common generation backbone \cite{qwen2026qwen35}. Original retrieval resources are retained when available; otherwise, systems receive a frozen 300-paper candidate set. Complete topics, configurations, prompts, and timeout rules are provided in \cref{app:das_bench}.

\noindent\textbf{\evalname{} metrics.}
\evalname{} evaluates publication-oriented surveys along four dimensions: Balanced Scholarly Citation Quality (BSC) measures citation grounding and distribution; Taxonomic Synthesis Quality (TSQ) measures literature coverage and global organization; Hierarchical Discourse Quality (HDQ) measures argumentation across sections and paragraphs; and Manuscript Assembly Reliability (MAR) measures reference integrity, visual integration, layout, and component completeness. Each dimension contains four criteria scored from 1 to 5, and the overall score is the mean of all 16 criteria. Multimodal LLM judges evaluate rendered manuscripts together with structured citation metadata, while domain experts independently rank method-blinded manuscripts. Complete rubrics and evaluation procedures are provided in \cref{app:das_eval}.

\begin{table*}[!t]
\centering
\small
\renewcommand{\arraystretch}{1.00}
\setlength{\tabcolsep}{1.0pt}
\resizebox{\textwidth}{!}{%
\begin{tabular}{c l ccccc ccccc ccccc ccccc c}
\toprule
&
\raisebox{-1.3ex}[0pt][0pt]{Method}
& \multicolumn{5}{c}{BSC$\uparrow$}
& \multicolumn{5}{c}{TSQ$\uparrow$}
& \multicolumn{5}{c}{HDQ$\uparrow$}
& \multicolumn{5}{c}{MAR$\uparrow$}
& \raisebox{-1.3ex}[0pt][0pt]{Total Avg.} \\
\cmidrule(lr){3-7}
\cmidrule(lr){8-12}
\cmidrule(lr){13-17}
\cmidrule(lr){18-22}
&
& Sup.
& Attr.
& MSyn.
& Bal.
& Avg.
& Cov.
& Bnd.
& Org.
& Ins.
& Avg.
& Aln.
& Prog.
& Spec.
& LSyn.
& Avg.
& Ref.
& Vis.
& Lay.
& Comp.
& Avg.
& \\
\midrule

& Human
& 3.87 & 3.90 & 3.53 & 4.07 & 3.84
& 4.23 & 4.23 & 4.40 & 4.30 & 4.29
& 4.47 & 4.07 & 4.37 & 4.07 & 4.24
& 5.00 & 5.00 & 5.00 & 5.00 & 5.00
& 4.34 \\

\midrule

\multicolumn{1}{c|}{\raisebox{-1.62\normalbaselineskip}[0pt][0pt]{\rotatebox[origin=c]{90}{\small General}}}
& Codex
& 3.00 & 3.43 & 2.73 & 2.03 & 2.80
& 3.00 & 3.00 & 3.00 & 2.97 & 2.99
& 3.43 & 2.43 & 3.07 & 2.07 & 2.75
& \textbf{5.00} & 1.77 & \textbf{5.00} & \textbf{5.00} & 4.19
& 3.18 \\

\multicolumn{1}{c|}{} & GPT DR
& 3.57 & 3.77 & 3.33 & 2.60 & 3.32
& 3.47 & 3.50 & 3.53 & 3.43 & 3.48
& 4.10 & 3.47 & 4.00 & 3.47 & 3.76
& \textbf{5.00} & 1.57 & \textbf{5.00} & \textbf{5.00} & 4.14
& 3.68 \\

\multicolumn{1}{c|}{} & Gemini DR
& 2.83 & 2.60 & 2.87 & 3.50 & 2.95
& 3.80 & 3.83 & 3.83 & 3.80 & 3.82
& 4.17 & \underline{3.97} & 4.17 & 3.97 & 4.07
& \underline{4.60} & \underline{4.87} & \underline{4.90} & \textbf{5.00} & \underline{4.84}
& 3.92 \\

\multicolumn{1}{c|}{} & Naive RAG
& \textbf{4.00} & \underline{4.00} & 3.33 & \underline{3.60} & 3.73
& \underline{3.97} & \textbf{4.30} & \underline{4.20} & 3.77 & \underline{4.06}
& \textbf{4.47} & 3.93 & \textbf{4.53} & 3.93 & \underline{4.22}
& \textbf{5.00} & 1.37 & \textbf{5.00} & \textbf{5.00} & 4.09
& \underline{4.03} \\

\midrule

\multicolumn{1}{c|}{\raisebox{-1.62\normalbaselineskip}[0pt][0pt]{\rotatebox[origin=c]{90}{\small Survey Gen.}}}
& AutoSurvey
& \textbf{4.00} & \underline{4.00} & 3.29 & \textbf{3.95} & \underline{3.81}
& \textbf{4.00} & 3.86 & 3.10 & \underline{4.00} & 3.74
& 4.10 & 3.14 & 4.24 & 3.29 & 3.69
& \textbf{5.00} & 1.29 & \textbf{5.00} & 3.38 & 3.67
& 3.73 \\

\multicolumn{1}{c|}{} & SurveyForge
& \textbf{4.00} & 3.90 & \underline{3.81} & 3.19 & 3.73
& \textbf{4.00} & \underline{4.00} & 3.24 & \underline{4.00} & 3.81
& 4.10 & 3.29 & 4.19 & 3.38 & 3.74
& \textbf{5.00} & 1.29 & \textbf{5.00} & \underline{4.05} & 3.83
& 3.78 \\

\multicolumn{1}{c|}{} & LiRA
& \underline{3.97} & 3.97 & 3.57 & 3.00 & 3.63
& \textbf{4.00} & 3.33 & 3.53 & \underline{4.00} & 3.72
& 3.97 & 3.90 & \underline{4.43} & 3.93 & 4.06
& 3.00 & 1.40 & 4.87 & 3.00 & 3.07
& 3.62 \\

\multicolumn{1}{c|}{} & Inter. Survey
& 3.10 & 3.00 & 1.93 & 2.97 & 2.75
& \textbf{4.00} & 3.70 & 3.83 & \underline{4.00} & 3.88
& 3.80 & 3.93 & 4.00 & \underline{4.00} & 3.93
& 4.13 & 4.77 & 4.83 & \textbf{5.00} & 4.68
& 3.81 \\

\midrule

& \textbf{\method{}}
& \textbf{4.00} & \textbf{4.03} & \textbf{4.00} & 3.37 & \textbf{3.85}
& \textbf{4.00} & \underline{4.00} & \textbf{4.37} & \textbf{4.50} & \textbf{4.22}
& \underline{4.27} & \textbf{4.13} & \textbf{4.53} & \textbf{4.17} & \textbf{4.28}
& \textbf{5.00} & \textbf{5.00} & \textbf{5.00} & \textbf{5.00} & \textbf{5.00}
& \textbf{4.34} \\

\bottomrule
\end{tabular}%
}
\caption{\textbf{System-level results on \benchmark{}.} All criteria use a 1--5 scale; Group Avg. and Total Avg. average four and 16 criteria, respectively. Best and second-best system scores are bold and underlined, while Human is reported for reference only. Full definitions and aggregation details are provided in \cref{app:das_eval}.}
\label{tab:main_results}
\end{table*}

\begin{table*}[!t]
\centering
\small
\renewcommand{\arraystretch}{1.08}
\setlength{\tabcolsep}{1.2pt}
\resizebox{\textwidth}{!}{%
\begin{tabular}{l ccccc ccccc ccccc ccccc c}
\toprule
\raisebox{-1.3ex}[0pt][0pt]{Variant}
& \multicolumn{5}{c}{BSC$\uparrow$}
& \multicolumn{5}{c}{TSQ$\uparrow$}
& \multicolumn{5}{c}{HDQ$\uparrow$}
& \multicolumn{5}{c}{MAR$\uparrow$}
& \raisebox{-1.3ex}[0pt][0pt]{Total Avg.} \\
\cmidrule(lr){2-6}
\cmidrule(lr){7-11}
\cmidrule(lr){12-16}
\cmidrule(lr){17-21}
& Sup.
& Attr.
& MSyn.
& Bal.
& Avg.
& Cov.
& Bnd.
& Org.
& Ins.
& Avg.
& Aln.
& Prog.
& Spec.
& LSyn.
& Avg.
& Ref.
& Vis.
& Lay.
& Comp.
& Avg.
& \\
\midrule

w/o Metadata
& 4.00 & 3.97 & 4.00 & 3.47 & 3.86
& 4.00 & 4.00 & 4.23 & 4.23 & 4.12
& 4.20 & 4.17 & 4.30 & 4.17 & 4.21
& 5.00 & 5.00 & 5.00 & 5.00 & 5.00
& 4.30 \\

w/o Taxonomy
& 4.00 & 4.03 & 4.00 & 3.70 & 3.93
& 4.03 & 4.03 & 4.27 & 4.30 & 4.16
& 4.23 & 4.23 & 4.47 & 4.27 & 4.30
& 5.00 & 5.00 & 5.00 & 5.00 & 5.00
& 4.35 \\

w/o Routing
& 4.00 & 4.00 & 4.00 & 3.03 & 3.76
& 4.00 & 4.00 & 4.17 & 4.23 & 4.10
& 4.17 & 4.07 & 4.37 & 4.07 & 4.17
& 5.00 & 5.00 & 5.00 & 5.00 & 5.00
& 4.26 \\

w/o Hier. Draft
& 4.00 & 4.03 & 3.97 & 3.07 & 3.77
& 4.07 & 4.07 & 4.40 & 4.33 & 4.22
& 4.47 & 4.40 & 4.60 & 4.10 & 4.39
& 5.00 & 5.00 & 5.00 & 5.00 & 5.00
& 4.34 \\

w/o Det. Val.\textsuperscript{\ensuremath{\dagger}}
& 4.00 & 4.00 & 3.97 & 3.13 & 3.78
& 4.10 & 4.10 & 4.40 & 4.40 & 4.25
& 3.97 & 3.90 & 4.13 & 3.93 & 3.98
& 5.00 & 5.00 & 5.00 & 5.00 & 5.00
& 4.25 \\

w/o Sem. Review
& 4.03 & 4.03 & 4.00 & 3.17 & 3.81
& 4.03 & 4.03 & 4.07 & 4.10 & 4.06
& 4.03 & 4.03 & 4.33 & 4.07 & 4.12
& 5.00 & 5.00 & 5.00 & 5.00 & 5.00
& 4.25 \\

\hline
\textbf{Full \method{}}
& 4.00 & 4.03 & 4.00 & 3.37 & 3.85
& 4.00 & 4.00 & 4.37 & 4.50 & 4.22
& 4.27 & 4.13 & 4.53 & 4.17 & 4.28
& 5.00 & 5.00 & 5.00 & 5.00 & 5.00
& 4.34 \\

\bottomrule
\end{tabular}%
}
\caption{\textbf{Ablation results on \benchmark{}.} Group Avg. and Total Avg. average four and 16 criteria, respectively. For the \textsuperscript{\ensuremath{\dagger}} variant, three noncompiling outputs received minimal syntax corrections only for MAR rendering.}
\label{tab:ablation_results}
\end{table*}

\section{Experiments}
\label{section:exp}
\subsection{Setup}
\noindent\textbf{Baselines.}
We compare \method{} with two groups of systems. General research and writing baselines include Codex, GPT Deep Research, Gemini Deep Research, and Naive RAG, which drafts directly from the shared paper metadata without explicit taxonomy planning or review \cite{openai2025codex,openai2025deepresearch,google2024deepresearch}. Automated survey baselines include AutoSurvey, SurveyForge, InteractiveSurvey, and LiRA \cite{wang2024autosurvey,yan2025surveyforge,wen2025interactivesurvey,go2026lira}. SurveyX exceeded the 12-hour limit and is reported only in the completion analysis \cite{liang2025surveyx}; DeepSurvey is excluded because no public implementation was available \cite{yang2026deepsurvey}. Detailed configurations and output statistics are provided in \cref{app:das_bench,app:completion_efficiency}.

\noindent\textbf{Evaluation protocol.}
Codex, GPT Deep Research, and Gemini Deep Research are evaluated using their official native configurations and retrieval capabilities. For Naive RAG and the reproducible automated survey generation systems, we preserve the released system workflows while using Qwen3.5-397B-FP8 as a common generation backbone \cite{qwen2026qwen35}. This protocol separates comparisons with native closed-source systems from controlled comparisons among reproducible methods.

\subsection{Main Results and Analysis}

\noindent\textbf{Overall quantitative comparison.}
\cref{tab:main_results} reports system-level performance across the four \evalname{} dimensions. AutoSurvey and SurveyForge are evaluated on the 21 CS topics supported by their original corpora, whereas the remaining completed systems are evaluated on all 30 topics. For comparison under identical topic coverage, \cref{app:matched_cs_results} reports all systems on the shared 21-topic subset. The matched comparison preserves the same ordering among generated systems. The Human row contains one publicly available academic survey matched to each \benchmark{} topic, with its sources and selection criteria provided in \cref{app:das_bench}.

Among the systems evaluated on all 30 topics, \method{} achieves the highest group averages in BSC, TSQ, HDQ, and MAR, yielding a Total Avg. of 4.34 compared with 4.03 for Naive RAG. Across the quantities reported in \cref{tab:main_results}, \method{} is best or tied for best on 18 of the 21 measures. Its clearest advantages occur in multi-reference synthesis, global organization, research insight, paragraph progression, local synthesis, and visual integration, consistent with the literature organization, hierarchical drafting, and manuscript finalization mechanisms described in \cref{sec:agentic_das}. The advantage is not uniform across all criteria: Naive RAG obtains the highest taxonomy boundary and multi-level goal alignment scores, while AutoSurvey performs best on citation balance.

\begin{figure*}[!t]
    \centering
    \includegraphics[width=\textwidth]{\arxivroot 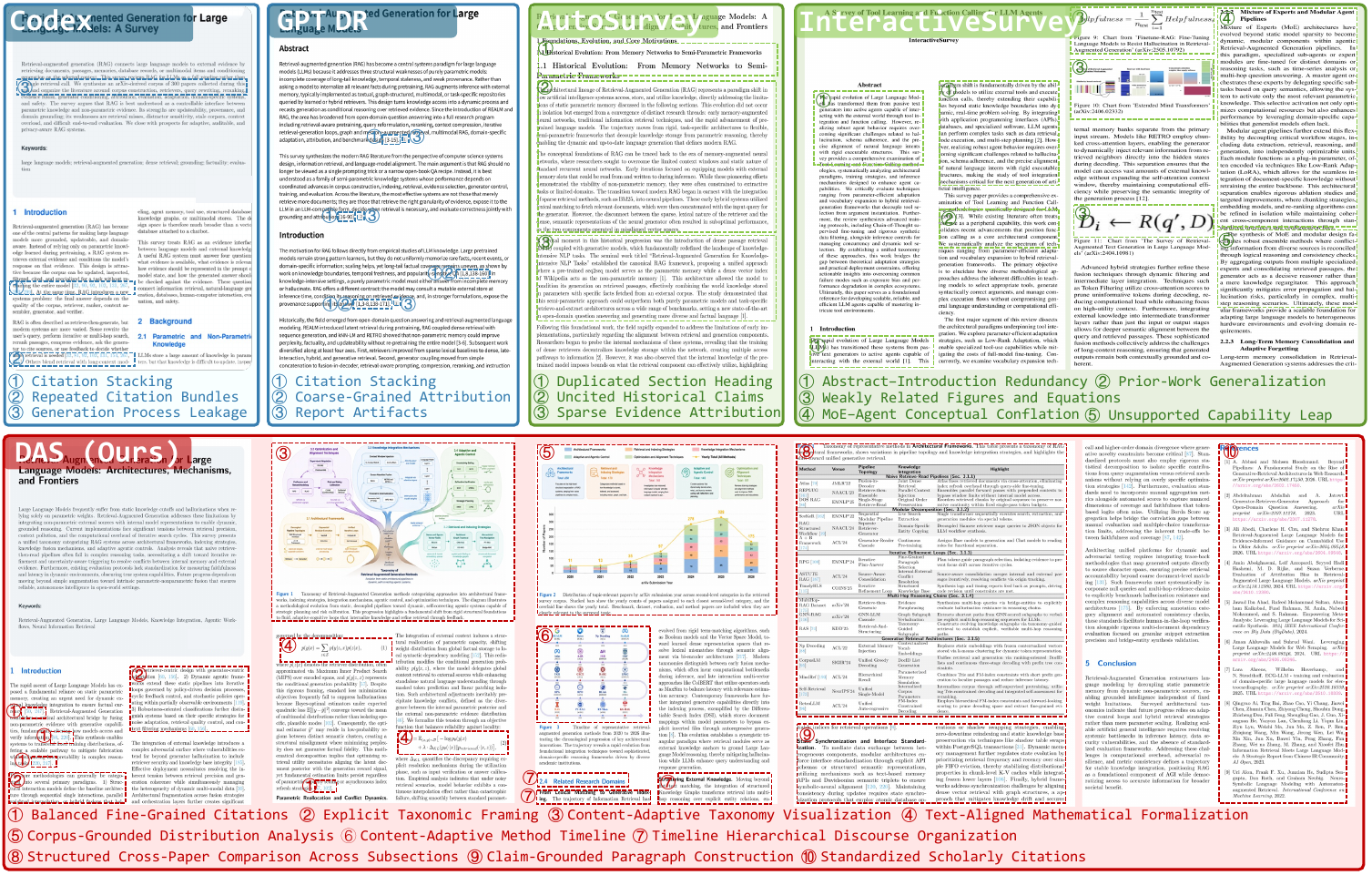}
    \caption{Qualitative comparison of survey artifacts from two general research systems (Codex and GPT DR), two automated survey generation systems (AutoSurvey and InteractiveSurvey), and \method{}. The figure presents representative first-page and interior-page views from the generated manuscripts.}
    \label{fig:das_exps}
\end{figure*}

\noindent\textbf{Qualitative manuscript case study.}
As shown in \cref{fig:das_exps}, the case study illustrates manuscript-level differences that are not fully captured by aggregate scores. Across the displayed baseline examples, the annotations identify specific instances of citation stacking, repeated citation bundles, coarse claim-to-reference attribution, duplicated headings, unsupported historical or capability statements, and weak coupling between figures, equations, and the surrounding discussion. For the same topic, the \method{} example maintains closer alignment among the taxonomy, section-level organization, claim-level citations, mathematical formulations, content-adaptive visualizations, and structured comparisons across papers. For conciseness, we present representative pages for each system as the remaining pages generally follow similar content and layout patterns. Complete generated manuscripts, additional qualitative comparisons, and the page selection protocol are included in the accompanying supplementary materials described in \cref{app:supplementary_outputs}.

\subsection{Ablation Studies}
\noindent\textbf{Ablation results on \benchmark{}.}
We ablate each core mechanism on all 30 topics while holding the remaining experimental conditions constant.
As shown in \cref{tab:ablation_results}, removing reverse paper-to-section routing decreases BSC, TSQ, and HDQ by 0.09, 0.12, and 0.11, respectively, while removing semantic review lowers both TSQ and HDQ by 0.16. Replacing structured paper representations with titles and abstracts reduces TSQ by 0.10 and HDQ by 0.07, indicating that richer metadata primarily benefits literature organization and discourse construction.

Taxonomy planning and hierarchical drafting exhibit cross-dimensional trade-offs. Removing taxonomy planning slightly increases BSC and HDQ but reduces TSQ, particularly global organization and research insight. Removing hierarchical drafting improves several discourse scores but lowers BSC, especially citation balance. Deterministic validation primarily improves execution reliability. Full \method{} compiles all 30 manuscripts, whereas the variant without it compiles 27 and reduces HDQ from 4.28 to 3.98. The three failed outputs received only minimal syntax corrections for MAR evaluation. Overall, reverse routing and semantic review provide the most consistent gains, while the remaining mechanisms contribute through dimension-specific improvements and manuscript reliability.

\Needspace{6\baselineskip}
\par\noindent\textbf{Repair policy ablation.}\par
\begin{wraptable}{r}{0.62\textwidth}
\vspace{-1.35\baselineskip}
\centering
\footnotesize
\renewcommand{\arraystretch}{1.08}
\setlength{\tabcolsep}{2.0pt}
\resizebox{\linewidth}{!}{%
\begin{tabular}{lccccc}
\toprule
Policy & BSC$\uparrow$ & TSQ$\uparrow$ & HDQ$\uparrow$ & RPR$\uparrow$ & RTok.$\downarrow$ \\
\midrule
Direct Edit Only       & 3.84 & 4.11 & 4.24 & 58.42\% & 0.95 \\
Paragraph Replan Only  & 3.87 & 4.21 & 4.38 & 53.69\% & 0.96 \\
Section Replan Only    & 3.81 & 4.13 & 4.23 & 45.13\% & 2.57 \\
w/o Semantic Review    & 3.81 & 4.06 & 4.12 & --      & 0.00 \\
\hline
\textbf{Full \method{}} & 3.85 & 4.22 & 4.28 & 74.59\% & 0.79 \\
\bottomrule
\end{tabular}%
}
\caption{Comparison of repair policies on \benchmark{} from identical pre-review states. RPR denotes the proportion of reviewed subsections that obtain PASS within the review budget, and RTok. denotes the review and repair tokens consumed per survey in millions.}
\label{tab:repair_policy_ablation}
\vspace{-0.35\baselineskip}
\end{wraptable}

To isolate the effect of repair scope, we run all policies from identical pre-review states under the same experimental settings. As shown in \cref{tab:repair_policy_ablation}, Full \method{} achieves the highest Review Pass Rate of 74.59\% and the lowest review cost among policies that perform semantic review, with an average of 0.79M tokens per survey. Although Paragraph Replan Only obtains slightly higher BSC and HDQ scores, it yields a lower pass rate and a higher review cost. Overall, the adaptive policy provides a favorable balance between review success and cost by matching the repair scope to the detected defect.

\subsection{Backbone Sensitivity Analysis}
We select five topics before running the sensitivity experiment: retrieval augmented
generation (003), dynamic 3D reconstruction (010), efficient LLM serving (015),
causal representation learning (020), and battery materials discovery (025). They
cover language systems, vision, infrastructure, machine learning, and a non-CS
scientific domain. Qwen3.5-397B-A17B-FP8, Qwen3.5-35B-A3B-FP8, and GPT-5.5 use the
same candidates, \method{} configuration, prompts, review budget, and evaluation
protocol. Table~\ref{tab:backbone_sensitivity} reports the frozen results on five topics;
no value is extrapolated to all 30 topics.

\begin{table*}[t]
\centering
\small
\renewcommand{\arraystretch}{1.0}
\setlength{\tabcolsep}{1.0pt}
\resizebox{\textwidth}{!}{%
\begin{tabular}{l ccccc ccccc ccccc ccccc c}
\toprule
\raisebox{-1.3ex}[0pt][0pt]{Backbone}
& \multicolumn{5}{c}{BSC$\uparrow$}
& \multicolumn{5}{c}{TSQ$\uparrow$}
& \multicolumn{5}{c}{HDQ$\uparrow$}
& \multicolumn{5}{c}{MAR$\uparrow$}
& \raisebox{-1.3ex}[0pt][0pt]{Total} \\
\cmidrule(lr){2-6}\cmidrule(lr){7-11}\cmidrule(lr){12-16}\cmidrule(lr){17-21}
& Sup. & Attr. & MSyn. & Bal. & Avg.
& Cov. & Bnd. & Org. & Ins. & Avg.
& Aln. & Prog. & Spec. & LSyn. & Avg.
& Ref. & Vis. & Lay. & Comp. & Avg. & \\
\midrule
GPT-5.5
&4.00&4.40&4.00&3.20&3.90
&4.40&4.20&4.80&5.00&4.60
&4.60&5.00&4.40&4.60&4.65
&5.00&5.00&5.00&5.00&5.00&4.54\\
Qwen3.5-35B-A3B-FP8
&3.60&3.60&3.60&3.00&3.45
&3.80&3.80&4.00&4.00&3.90
&3.80&3.20&4.00&3.00&3.50
&5.00&5.00&5.00&5.00&5.00&3.96\\
\midrule
Qwen3.5-397B-A17B-FP8
&4.00&4.00&4.00&3.00&3.75
&4.00&4.00&4.60&5.00&4.40
&4.20&4.20&4.60&4.40&4.35
&5.00&5.00&5.00&5.00&5.00&4.38\\
\bottomrule
\end{tabular}%
}
\caption{Backbone sensitivity on five stratified topics. All rows use identical \method{} inputs and settings. Group averages summarize four submetrics, and Total is the mean of all 16 submetrics.}
\label{tab:backbone_sensitivity}
\end{table*}

GPT-5.5 achieves the highest Total score, improving over the default backbone by
0.16 points, with the largest gains in TSQ and HDQ. The 35B model reduces Total by
0.42 points but retains a TSQ average of 3.90 and complete manuscript artifacts.
These results show that backbone capacity affects analytical quality, while the
fixed literature organization, planning, validation, and assembly mechanisms remain
effective across model scales. GPT-5.5 offers the strongest quality but requires a
proprietary API, whereas the 35B model provides a lower resource option. The default
397B backbone therefore offers a practical balance between output quality and
inference cost.

\FloatBarrier
\subsection{Human Evaluation and Cross-Judge Robustness}
\noindent\textbf{Expert evaluation.}
Three domain experts independently ranked method-blinded manuscripts according to literature support, organization, analytical writing, and completeness. Under majority voting, \method{} was preferred to Naive RAG on 27 of 30 topics and to AutoSurvey on 19 of 21 shared CS topics, ranking first on 18 of these 21 topics. The experts agreed unanimously on 63 of 72 pairwise comparisons. Protocol and complete results are provided in \cref{app:human_judge}.

\noindent\textbf{Cross-judge robustness.}
Kimi K2.6 re-evaluation yields a moderate correlation with Qwen3.5 ($\rho=0.507$, MAE $=0.630$) while preserving the ordering \method{} $>$ Naive RAG $>$ AutoSurvey. The judges agree on 48 of 63 CS comparisons and both rank \method{} above Naive RAG on the non-CS subset, although local agreement is lower. Thus, the overall ranking is stable, while fine-grained scores remain judge-sensitive. Detailed results are provided in \cref{app:human_judge}.

\section{Conclusion}
We presented \method{}, a stateful agentic framework that formulates publication-oriented survey generation as a closed-loop manuscript construction process. Building on the dynamically updated \dataset{} literature metadata lake, \method{} integrates candidate-grounded literature organization, hierarchical claim and citation planning, scoped semantic repair, and deterministic finalization to maintain consistency from candidate papers to the compiled manuscript. We also introduced \benchmark{} for evaluating literature support, taxonomic organization, hierarchical discourse, and manuscript reliability across 30 topics. Experimental and expert evaluations show that \method{} achieves the strongest overall performance among the compared systems, while the ablations clarify the distinct contributions and trade-offs of its core mechanisms. These findings indicate that manuscript states and repair provide an effective foundation for constructing reliable academic surveys at scale.

\bibliography{\arxivroot aaai2027,\arxivroot supplementary_human}

\clearpage
\appendix
\begin{center}
    {\huge\sffamily\bfseries\color{aprilblue} Appendix}
\end{center}
\setcounter{secnumdepth}{2}
\renewcommand\thefigure{A\arabic{figure}}
\renewcommand\thetable{A\arabic{table}}
\renewcommand\theequation{A\arabic{equation}}
\setcounter{equation}{0}
\setcounter{table}{0}
\setcounter{figure}{0}
\label{appendix}

The appendix documents the supplementary website, the construction and audit of
\dataset{}, the implementation of \method{}, the benchmark and evaluation
protocols, additional results, human assessment, and the planned release.

\begin{itemize}
\item[-] \cref{app:supplementary_outputs} describes the static website and the page selection protocol used for qualitative comparison.
\item[-] \cref{app:das2m} documents the construction and quality assurance of \dataset{}.
\item[-] \cref{app:das_implementation} provides the implementation details of \method{}.
\item[-] \cref{app:das_bench} presents \benchmark{}, baseline configurations, and the common generation protocol.
\item[-] \cref{app:das_eval} defines the complete \evalname{} rubrics and evaluation procedure.
\item[-] \cref{app:additional_results} reports additional experimental results and analyses.
\item[-] \cref{app:human_judge} details the human evaluation and cross-judge robustness analysis.
\item[-] \cref{app:reproducibility} summarizes the release plan, reproducibility information, and limitations.
\end{itemize}

\section{Supplementary Website and Artifact Index}
\label{app:supplementary_outputs}

The website is self-contained. From the root of the Code and Data Supplement,
readers can open \texttt{website/index.html} in a modern browser; no network
connection, package installation, or local server is required. The website provides
an interactive visualization of the \method{} construction process, showing how
literature metadata, taxonomy planning, paper routing, hierarchical drafting,
review actions, and final artifacts are connected throughout the workflow. It also
supports a direct visual comparison between \method{} and representative baselines
on the topic used in Figure~3 of the main paper. Readers can switch among the
complete surveys while preserving the same comparison target. In addition, the
gallery presents three complete \method{} surveys from different topics.

For Figure~3 of the main paper, we use the first page to show the manuscript identity and
overall presentation, followed by representative interior pages that expose method
organization, citation use, and visual integration. The same rule is applied to all
systems before annotations are added. Baseline manuscripts whose interior pages
repeat the same text and layout pattern are represented by the first eligible
interior page. Complete manuscripts remain available on the website so that the
selection can be independently inspected.

The \texttt{website/assets/} directory stores the display assets, and
\texttt{website/examples/} contains the complete benchmark and \method{} manuscript
artifacts used by the gallery. All quantitative results are computed from the
archived evaluation inputs described in \cref{app:das_eval}.

\section{\dataset{} Construction and Quality Assurance}
\label{app:das2m}

\subsection{Corpus Collection and Statistics}
\dataset{} is the persistent literature metadata resource used by \method{}. The
name refers to its acquisition scale: we collected approximately 2 million arXiv
PDF files released between January 2020 and June 2026 across all subject categories
from the arXiv corpus available through Google's Kaggle platform. When multiple
versions of the same arXiv paper were available, only the latest version was retained. After
deduplication and removal of corrupted or unparsable files, 1.53 million papers
remained in the experimental resource. \dataset{} is dynamically maintained through
the same ingestion and processing pipeline as new papers become available.

\subsection{Document Parsing and Metadata Extraction}
PDFs are parsed offline with the MinerU pipeline and PDF-Extract-Kit-1.0 model
assets. We use automatic parsing for English documents, enable table recognition,
disable formula recognition, and retain the reconstructed Markdown and content
list. Each paper has a 600-second parsing budget. Before metadata extraction, the
pipeline removes control characters, HTML tables, embedded images, and trailing
reference or appendix material detected near the end of the document. Cleaned
inputs longer than 400,000 characters are explicitly marked as skipped; inputs
between 300,000 and 400,000 characters are truncated to 300,000 characters.

Qwen3.5-397B-A17B-FP8 extracts the metadata with thinking disabled, five concurrent
workers, temperature 0.7, top-$p$ 0.8, top-$k$ 20, presence penalty 1.5, and a
4,096-token output limit. A request is retried up to ten times with increasing
delays. The parser requires one JSON object, attempts conservative JSON repair when
needed, and writes the canonical source path programmatically. Unmentioned evidence
dependent fields are represented by \texttt{null}.

\begin{lstlisting}[caption={Condensed instruction for survey metadata extraction. The complete output schema is given in \cref{tab:metadata_schema}.},label={lst:metadata_prompt},captionpos=t,frame=tb,numbers=none,xleftmargin=0pt]
Role: Read the supplied academic paper and extract a survey-oriented record.

Requirements:
1. Return one valid JSON object and no surrounding text.
2. Write all values in English.
3. Use only information supported by the paper. Use null when the paper does not provide a field.
4. Record at most the first three full author names; do not use "et al."
5. Use an integer publication year and YYYY-MM when the month is available.
6. Preserve formulas with valid JSON escaping.
7. Use concise technical descriptions and omit promotional language.
8. Follow exactly the groups, fields, and value types in the metadata schema table.

Input paper:
{parsed_markdown}
\end{lstlisting}

\subsection{Survey-Oriented Metadata Schema}
The schema groups information by its survey function rather than storing a single
free form summary. \cref{tab:metadata_schema} lists all 25 fields. For normal
extraction outputs, the eight groups and 25 field keys are fixed; values may be
\texttt{null} when unsupported, while the source pointer is injected
programmatically.

\FloatBarrier
\begin{table}[!ht]
\centering
\small
\setlength{\tabcolsep}{3.4pt}
\begin{tabular}{@{}p{2.3cm}p{6.6cm}p{3.0cm}p{3.8cm}@{}}
\toprule
Group & Fields & Value type & Principal use \\
\midrule
Basic information & title; authors; publication year; publication date; source path
& string / list / integer / null & identity and source access \\
Categorization & task category; keywords
& string / list & retrieval and taxonomy \\
Technical attributes & pipeline; learning paradigm; knowledge source; backbone
& string or null & retrieval, taxonomy, routing \\
Resource information & code availability; repository; project page
& Boolean / string / null & resource reporting \\
Method details & method name; architecture description; innovations
& string / list & routing and method drafting \\
Dataset information & new dataset flag; dataset name; evaluation datasets
& Boolean / string / list / null & comparison and evaluation synthesis \\
Research logic & motivation; key insights; conclusion
& string or null & taxonomy, challenges, prospects \\
Evaluation & main results; limitations
& object list / string / null & claim planning and critical analysis \\
\bottomrule
\end{tabular}
\caption{Survey-oriented metadata schema. Container groups and field keys are fixed; values may be null when evidence is unavailable.}
\label{tab:metadata_schema}
\end{table}
\FloatBarrier

Different stages receive restricted views of this record. Retrieval materializes
lexical and semantic text from topical, technical, methodological, and evaluation
fields. Taxonomy planning additionally observes motivations, insights, and
limitations. Routing adds method architecture, innovations, dataset information,
and conclusions. Paragraph planning uses a slim view containing identity, method
tags, innovation, and core insight, whereas claim planning loads the full record
only for papers assigned to that paragraph. Prose realization receives the claim
and citation plan, optional source evidence, and preceding text, but not the raw
metadata record.

\begin{lstlisting}[caption={Illustrative schema example. Long text fields are shortened for presentation.},label={lst:metadata_example},captionpos=t,frame=tb,numbers=none,xleftmargin=0pt]
{
  "basic_info": {"title": "Example Paper", "authors": ["Author A", "Author B"], "publication_year": 2020, "publication_date": "2020-06", "file_path": "<source_pointer>"},
  "categorization": {"task_category": "Biometric Identification", "keywords": ["Continual Learning", "Representation Learning"]},
  "technical_attributes": {"pipeline_type": "Continual representation learning", "learning_paradigm": "Supervised continual learning", "knowledge_source": "Sequential biometric tasks", "backbone_model": null},
  "resource_info": {"has_code": false, "github_url": null, "project_page": null},
  "method_details": {"method_name": "Continual representation method", "architecture_description": "Updates a shared representation across sequential identification tasks.", "innovations": ["Preserves transferable identity features across tasks"]},
  "dataset_info": {"contributed_new_dataset": false, "new_dataset_name": null, "datasets_used_for_eval": ["Biometric benchmarks"]},
  "research_logic": {"motivation": "Sequential enrollment changes the identity distribution.", "key_insights": "Continual updates can preserve discriminative representations.", "conclusion": "Representation retention improves sequential identification."},
  "evaluation": {"main_results": [{"dataset": "Biometric benchmark", "metric": "Identification accuracy", "value": "reported in source", "comparison": null}], "limitations": "Performance depends on the sequence of tasks."}
}
\end{lstlisting}
\FloatBarrier

\subsection{Metadata Integrity Audit}
We rebuild a 200-record audit sample from the current frozen directories using
random sampling stratified by year with seed \texttt{20270731}. The allocation for
2020 through 2026 is 25, 26, 26, 29, 34, 40, and 20 records, respectively. All 200
records are valid JSON, contain the eight metadata groups and 25 expected fields
with valid types, and resolve to parsed source documents. Because \dataset{} is
continually extended, a seed alone cannot reproduce a sample after the underlying
directory changes. We therefore retain the explicit sampled identifiers with the
audit release. A normalized exact lexical check recovers 195 of 200 titles and 503
of 506 eligible author surnames from the source documents; this check tests source
traceability and does not assess semantic correctness.
\FloatBarrier
\begin{table}[!ht]
\centering
\small
\setlength{\tabcolsep}{4.2pt}
\begin{tabular}{lrr}
\toprule
Audit item & Count & Rate \\
\midrule
Valid JSON & 200/200 & 100.0\% \\
Normal extraction & 200/200 & 100.0\% \\
Eight groups present & 200/200 & 100.0\% \\
All 25 fields present & 200/200 & 100.0\% \\
Valid field types & 200/200 & 100.0\% \\
Resolvable source pointer & 200/200 & 100.0\% \\
Title lexical recovery & 195/200 & 97.5\% \\
Author-surname recovery & 503/506 & 99.4\% \\
\bottomrule
\end{tabular}
\caption{Structural integrity and traceability audit of 200 records sampled with a fixed seed and stratified by year.}
\label{tab:metadata_audit}
\end{table}
\FloatBarrier

We further draw 20 records from this sample with seed \texttt{20270801}, preserving
the yearly allocation at 2, 3, 3, 3, 3, 4, and 2 records. Two isolated Codex model
instances independently compare each metadata record with its parsed source. Neither
model auditor receives the other audit output or the result of the preceding audit.
For each paper, both independent model-based audits examine method or mechanism,
innovations, evaluation results, limitations, and research logic. Correctness
requires every material statement in a domain to be supported by the source without
contradiction. Central completeness requires the metadata to cover the principal
source content for that domain, without requiring every minor detail. Unsupported
content records whether any material statement lacks source support or conflicts
with a reported value or conclusion.

Audit A judges 93 of 100 domains correct, all 100 centrally complete, and seven as
containing unsupported content. Audit B judges 97 domains correct, 99 centrally
complete, and three as containing unsupported content. Treating a domain as aligned
only when it is correct, centrally complete, and free of unsupported content, the two
audits agree on 92 of 100 domain decisions, yielding 92.0\% raw agreement. The
outcomes comprise 91 positive agreements, one negative agreement, and eight
disagreements. Because aligned domains dominate the sample, we report these
underlying counts together with the aggregate agreement. Across the three binary
labels separately, raw agreement is 94.3\% (283/300). A third model adjudication
returns to the source passages for the eight disputed domain
decisions, retaining five errors and rejecting three. We also apply a deterministic
arithmetic check to quantitative comparisons; it identifies one additional error
that both model auditors missed.

\begin{table}[t]
\centering
\small
\setlength{\tabcolsep}{4.2pt}
\begin{tabular}{lrrrr}
\toprule
Semantic domain & Judgments & Correct & Complete & Unsupported \\
\midrule
Method or mechanism & 20 & 19/20 & 20/20 & 1/20 \\
Innovations & 20 & 20/20 & 20/20 & 0/20 \\
Evaluation results & 20 & 16/20 & 19/20 & 4/20 \\
Limitations & 20 & 19/20 & 20/20 & 1/20 \\
Research logic & 20 & 19/20 & 20/20 & 1/20 \\
\midrule
Overall & 100 & 93/100 & 99/100 & 7/100 \\
\bottomrule
\end{tabular}
\caption{Adjudicated results of two independent model-based source audits on 20 metadata records. Complete denotes coverage of the central source content, and unsupported denotes at least one material statement without source support.}
\label{tab:metadata_semantic_audit}
\end{table}

The adjudicated errors comprise an incorrect order-of-magnitude comparison in
arXiv:2011.03522; unsupported or overgeneralized evaluation statements in
arXiv:2101.04405, 2302.03056, and 2405.17794; an untested applicability boundary in
arXiv:2502.09388; and incorrect input and result descriptions in arXiv:2504.19141.
For the first case, 24 seconds relative to approximately $10^4$ seconds is described
as a four-order reduction, whereas the reported values imply about $417\times$, or
2.62 orders of magnitude. These findings indicate high source alignment in the
sample while also showing why metadata should remain a planning representation with
recoverable source pointers, rather than an error-free substitute for source papers.

\begin{table}[!t]
\centering
\small
\setlength{\tabcolsep}{4.0pt}
\begin{tabular}{lp{0.78\textwidth}}
\toprule
Year & Audited arXiv identifiers \\
\midrule
2020 & 2011.03522; 2010.14050 \\
2021 & 2101.04405; 2109.02987; 2112.09319 \\
2022 & 2208.04914; 2205.13391; 2210.12511 \\
2023 & 2311.18600; 2312.16353; 2302.03056 \\
2024 & 2405.17794; 2412.05256; 2409.18525 \\
2025 & 2502.09388; 2504.19141; 2509.16304; 2511.06544 \\
2026 & 2605.31271; 2605.26283 \\
\bottomrule
\end{tabular}
\caption{Identifiers in the 20-record semantic audit.}
\label{tab:metadata_semantic_manifest}
\end{table}

\subsection{Index Construction and Maintenance}
The retrieval store materializes two text views. The lexical view emphasizes title,
task, keywords, method name, datasets, technical attributes, innovations, and main
results. The semantic view additionally includes architecture, motivation,
insights, conclusion, and limitations. We encode the latter with
Qwen3-Embedding-0.6B into 1,024-dimensional vectors, using batches of 64 and a
maximum input length of 8,192 tokens. LanceDB provides a text index for the
lexical view and an IVF-PQ cosine index with 256 partitions and 64 subvectors for
the semantic view. Online retrieval executes lexical and dense queries and merges
their rankings by weighted reciprocal rank fusion. For paper $i$ and query $q$, the
fusion score is
\begin{equation}
\operatorname{RRF}(i\mid q)
=\sum_{r\in\mathcal{R}(q)}
\frac{w_r^{\mathrm{route}}w_r^{\mathrm{query}}}
{60+\operatorname{rank}_r(i)},
\label{eq:weighted_rrf}
\end{equation}
where $\mathcal{R}(q)$ contains the lexical and dense retrieval routes,
$w_r^{\mathrm{route}}$ and $w_r^{\mathrm{query}}$ are their route and query
weights, and $\operatorname{rank}_r(i)$ is the rank of paper $i$ returned by route
$r$. Papers are ordered by the summed score after duplicate identifiers are merged.

The monthly maintenance job reruns acquisition, parsing, metadata extraction, and
append mode index updates for newly available papers. Each yearly table and the
merged global table reject duplicate base arXiv identifiers. This procedure keeps
the searchable snapshot current with newly ingested papers; it does not imply an
automatic replacement of every previously ingested record when arXiv releases a
newer version.

\section{Implementation Details of \method{}}
\label{app:das_implementation}
This section documents the agent roles, state interfaces, planning and routing
procedures, validation rules, visual generation process, model settings, and
execution configuration cited as Section C in the main paper. The corresponding
executable prompts, structured parsers, configuration files, stage implementations,
and visual generation code are provided in the \texttt{code/} directory of the Code
and Data Supplement. The descriptions below and these versioned artifacts jointly
supply the cited implementation material.

\subsection{Agent Roles and State Interfaces}
\method{} exposes to each role only the state view required for its decision. LLM
outputs are parsed into a specified schema before they are committed, whereas
retrieval, deterministic checking, action dispatch, and compilation remain fixed
operators. Review feedback reopens only the affected writing state. This separation
prevents an unconstrained response from silently altering upstream literature or
organization states.

\begin{table*}[t]
\centering
\small
\setlength{\tabcolsep}{3.2pt}
\begin{tabular}{@{}p{2.4cm}p{4.0cm}p{5.2cm}p{4.1cm}@{}}
\toprule
Role or operator & State view & Decision or operation & Updated object \\
\midrule
Query Planner & topic and retrieval configuration & lexical, semantic, and background queries & $\mathcal{Q}_t$ \\
Hybrid Retriever & $\mathcal{Q}_t$ and offline indexes & weighted rank fusion & $\mathcal{C}_t$ \\
Taxonomy Planner & organization views of $\mathcal{C}_t$ & typed rooted taxonomy & $\mathcal{T}_t$ \\
Paper Router & $\mathcal{T}_t$ and batched metadata & zero to three target nodes per paper & $\mathcal{R}_t$ \\
Paragraph Planner & node role, routed papers, slim metadata & Route A draft or Route B paragraph plan & $d_{s,1}^{(k)}$ or $\mathcal{P}_{s}^{(k)}$ \\
Claim Planner & paragraph theme and assigned full metadata & writing points, citation groups, evidence request & $\mathcal{W}_{s,j}^{(k)}$ \\
Drafter & point plan, evidence, preceding text & paragraph prose & $d_{s,j}^{(k)}$ \\
Deterministic Validator & paragraph and admissible identifiers & normalization or revision instructions & validated paragraph candidate \\
Reviewer & complete subsection and section objective & repair action and critique & local $\mathcal{S}^{\mathrm{write}}_t$ \\
Finalizer & accepted text and artifact specifications & figures, tables, bibliography, and source assembly & $\mathcal{S}^{\mathrm{final}}_t$ \\
\bottomrule
\end{tabular}
\caption{Roles and fixed operators in the stateful \method{} workflow.}
\label{tab:agent_roles}
\end{table*}

\subsection{Taxonomy Planning and Paper Routing}
The Taxonomy Planner observes the organization view of the top 300 candidates. It
produces a rooted hierarchy whose nodes contain a stable identifier, title,
description, and a field determined by its role. Analytical nodes provide key questions,
reflective nodes provide strategic perspectives, and navigational nodes organize
their descendants. Deterministic checks enforce a valid hierarchy, required fields,
identifiers, numbering, and the expected Methods organization. They do not claim to
verify semantic coverage or disjointness.

The Paper Router processes ten papers at a time. For each paper, it compares the relevant
metadata facets with eligible taxonomy nodes and returns at most three targets;
empty assignment is valid for a weakly aligned candidate. Technical mechanisms and
innovations inform method sections, motivations and limitations inform challenge
sections, and conclusions and insights inform prospect sections. Invalid paper IDs,
unknown nodes, and duplicate targets are rejected. The organization stage permits
up to five attempts and retains the valid result with the highest assignment
coverage, using 60\% as the target coverage threshold. The resulting relation
defines the literature supplied to paragraph and citation planning; it is not
described as an independent global citation firewall.

\subsection{Hierarchical Planning and Drafting}
For each taxonomy node, the Paragraph Planner reads its role, structural position,
routed papers, and slim metadata. Route A directly drafts a concise overview for a
navigational node or a node whose assigned literature does not support detailed
analysis. Route B constructs an ordered paragraph plan whose entries specify a
theme, argumentative responsibility, target length, and supporting paper IDs.
Introduction, discussion, challenge, prospect, and conclusion nodes follow role
specific eligibility constraints rather than an unconstrained route choice.

For each Route B paragraph, the Claim Planner loads full metadata only for assigned
papers and creates a sequence of claim like writing points with citation groups.
An unsupported point must use \texttt{Cite: None}. If an indispensable detail is
absent, a focused request may access the parsed source through the stored pointer;
the budget is one request per paragraph and two per section. The Drafter then
receives the section objective, paragraph theme, point plan, optional evidence, and
validated preceding paragraphs. It does not receive the raw metadata collection.
This ordering commits paper selection and evidence scope before prose realization.

\subsection{Deterministic Validation and Scoped Repair}
After each paragraph is drafted or revised, a rule checker examines citation command
syntax, candidate identifier validity, required citations from the point plan,
adjacent duplicate citations, paragraph boundaries, unresolved placeholders,
section formatting, bare paper IDs, language purity, LaTeX special characters,
display mathematics, braces, and environments. Safe formatting violations are
normalized directly; remaining violations become explicit revision instructions.
Each paragraph has at most six draft and check attempts. Fatal compilation or
structural violations block commitment, while the bounded fallback may retain the
latest candidate if only nonfatal issues remain.

The Reviewer observes a complete Route B subsection, paragraph IDs, the
taxonomy objective, academic tasks, and the section role. It assesses technical
relevance, paragraph progression, cross paragraph redundancy, alignment with the
section objective, and citation coverage of central claims. It does not read source
papers and therefore is not used to claim source level entailment verification. The
reviewer is instructed to choose the smallest sufficient repair scope:
\texttt{PASS}, \texttt{DIRECT\_EDIT}, \texttt{REPLAN\_PARAGRAPH}, or
\texttt{REPLAN\_SECTION}. Revised components return through deterministic checking.
The review loop has at most three rounds; if the budget is exhausted or decisions
cannot be parsed, the latest complete subsection is retained.

\subsection{Visual Generation and Manuscript Assembly}
After writing, \method{} constructs four types of visual artifacts. A taxonomy
diagram derives its structure from method tables, section descriptions, and
representative methods. A distribution plot assigns up to 1,000 retrieved papers to
the closest second level method categories and counts them by arXiv submission year.
A timeline selects representative methods across years from the generated method
tables. Each second level Methods section may also receive an explanatory figure
whose specification is conditioned on its description, child subsections, and key
questions. These specifications are sent to the image model; they are not copied
from figures in the cited papers.

The integration stage inserts a concise \texttt{\textbackslash cref} sentence at the
corresponding section, verifies that required labels are referenced, and preserves
existing citations and labels. Bibliography generation extracts arXiv IDs actually
used in the final TeX, queries Semantic Scholar for venue, journal, and DOI records,
falls back to arXiv and local metadata, and sanitizes BibTeX. The Finalizer orders
section inputs and assets, checks referenced files, and compiles with
\texttt{latexmk -halt-on-error}. A successful PDF and the associated TeX, BibTeX,
tables, figures, and logs constitute the terminal artifact state.

\subsection{Models, Hyperparameters, and Resources}
\begin{table*}[t]
\centering
\small
\setlength{\tabcolsep}{4pt}
\begin{tabular}{lll}
\toprule
Component & Setting & Value \\
\midrule
Generation & backbone & Qwen3.5-397B-A17B-FP8 \\
Retrieval & embedding model & Qwen3-Embedding-0.6B \\
Retrieval & candidate budget / taxonomy budget & 1,000 / 300 \\
Thinking calls & output, temperature, top-$p$, top-$k$ & 32,768; 0.6; 0.95; 20 \\
Other LLM calls & output, temperature, top-$p$, top-$k$ & 32,768; 0.7; 0.8; 20 \\
Evidence access & paragraph / section budget & 1 / 2 \\
Workers & routing / main writing / visual integration & 5 / 4 / 5 \\
Workers & trend / table / T2I & 5 / 5 / 10 \\
Review & subsection rounds & 3 \\
Length settings & supported / evaluated & short, medium, long / medium \\
\bottomrule
\end{tabular}
\caption{Principal \method{} settings used for \benchmark{}.}
\label{tab:das_settings}
\end{table*}

Query planning and taxonomy construction use thinking mode; routing and the main
drafting calls use the nonthinking configuration. Schema validation uses up to three
stage attempts, while transient API failures are handled by the shared client retry
policy. We report logical concurrency because the model is served through an API;
the endpoint deployment topology is not used as an experimental variable.

\noindent\textbf{Length settings and execution order.}
The short, medium, and long settings change the requested section and paragraph
budgets while preserving the same state interfaces and validation logic. The
reported experiments use the medium setting. Online construction proceeds through
retrieval, taxonomy and routing, writing with section figure planning, trend
analysis, bibliography generation, method tables, taxonomy and timeline planning,
image generation, visual reference integration, LaTeX assembly, and PDF
compilation. Runtime and output statistics are reported in
\cref{app:completion_efficiency}.

\subsection{Prompt and Parser Release}
\label{app:complete_role_prompts}
The state interfaces, decision schemas, validation rules, and principal model
settings are specified in the preceding subsections. The complete runtime prompt
templates, structured parsers, and visual specification templates are archived in
the accompanying Code and Data Supplement under versioned role directories. Each
artifact records its expected input fields, output schema, and corresponding
configuration. This separation keeps the appendix focused on the scientific design
while preserving the exact executable materials required for reproduction.

\section{\benchmark{} and Generation Protocol}
\label{app:das_bench}
This section documents the complete topic set, common generation task, candidate
literature resources, baseline configurations, human reference surveys, and timeout
protocol cited as Section D in the main paper.

\subsection{Topic Set and Selection Procedure}
\benchmark{} contains 21 computer science topics and nine interdisciplinary
topics. We selected topics that support a substantial recent literature, admit
multiple technical families, and require organization beyond a single method
summary. The set spans language models, information retrieval, multimodal learning,
vision, robotics, systems, security, scientific discovery, biomedicine, materials,
climate, earth observation, finance, and uncertainty. Topic wording was fixed before
generation and evaluation.

\begin{table*}[t]
\centering
\small
\setlength{\tabcolsep}{3.5pt}
\begin{tabular}{clp{12.1cm}}
\toprule
ID & Domain & Survey topic \\
\midrule
001 & CS & Tool Learning and Function Calling for LLM Agents \\
002 & CS & Memory and Long-Context Mechanisms for Long-Horizon LLM Agents \\
003 & CS & Retrieval-Augmented Generation for Large Language Models \\
004 & CS & Planning and Self-Reflection in Large Language Model Reasoning \\
005 & CS & Prompt Injection and Tool-Use Security in LLM Agents \\
006 & CS & Program Repair and Automated Debugging with Code LLMs \\
007 & CS & Multimodal Retrieval-Augmented Generation for Chart and Document Understanding \\
008 & CS & Vision-Language Models for Embodied Reasoning \\
009 & CS & Diffusion and Flow-Based Models for Controllable Image Generation \\
010 & CS & Gaussian Splatting and Neural Rendering for Dynamic 3D Scene Reconstruction \\
011 & CS & Multi-Sensor Fusion for Autonomous Driving Perception \\
012 & CS & Continual Learning and Model Editing for Foundation Models \\
013 & CS & Offline and Preference-Based Reinforcement Learning for Robotics \\
014 & CS & Graph Neural Networks and Graph Foundation Models for Scientific Discovery \\
015 & CS & Efficient LLM Serving with KV Cache, Speculative Decoding, and Quantization \\
016 & CS & Vector Databases and Retrieval Systems for Large-Scale AI Applications \\
017 & CS & Privacy-Preserving Machine Learning with Federated Learning and Differential Privacy \\
018 & CS & AI Software Supply-Chain Security and Vulnerability Detection \\
019 & CS & Human-AI Collaboration in Scientific Writing and Research Workflows \\
020 & CS & Causal Representation Learning and Causal Discovery in Deep Learning \\
021 & CS & Large Language Models for Generative Recommendation and User Behavior Modeling \\
022 & non-CS & AI-Driven Protein Design with Diffusion and Language Models \\
023 & non-CS & Single-Cell Foundation Models for Cell Type Annotation and Perturbation Prediction \\
024 & non-CS & Radiomics and Deep Learning for Tumor Diagnosis and Prognosis \\
025 & non-CS & Machine Learning for Solid-State Battery Materials Discovery \\
026 & non-CS & Machine Learning for Electrocatalyst Discovery in Energy Conversion \\
027 & non-CS & Deep Learning for Extreme Weather Forecasting \\
028 & non-CS & Foundation Models for Satellite Earth Observation \\
029 & non-CS & Deep Learning for Financial Risk Modeling under Uncertainty \\
030 & non-CS & Bayesian Deep Learning for Uncertainty Quantification \\
\bottomrule
\end{tabular}
\caption{The 30 topics in \benchmark{}.}
\label{tab:bench_topics}
\end{table*}

\subsection{Generation Task and Common Output Requirements}
All systems are asked to produce a complete English academic survey grounded in
scholarly publications. The target of approximately 300 references is stated only
when the interface supports such a request and is not treated as a hard constraint
for every baseline. Closed systems retain their native search, planning, and writing
behavior. Reproducible systems retain their released workflows and receive the
resources specified in \cref{app:candidate_resources}.

\begin{table}[t]
\centering
\small
\setlength{\tabcolsep}{4pt}
\begin{tabularx}{\textwidth}{@{}p{2.8cm}X@{}}
\toprule
Interface & Task-specific requirements \\
\midrule
GPT and Gemini DR & Native literature search and report construction; scholarly sources, numbered citations, complete survey structure, and self-contained output. \\
Codex & The same scholarly survey objective, together with BibTeX generation, LaTeX compilation, citation mapping, and final PDF delivery. \\
Naive RAG & Direct generation from the frozen 300 metadata records; numbered citations must resolve to the supplied records, and no external literature may be introduced. \\
\bottomrule
\end{tabularx}
\caption{Interface-specific additions to the common survey generation task.}
\label{tab:common_generation_task}
\end{table}

The complete verbatim task files, including topic and path placeholders, are included
in the Code and Data Supplement. They are versioned separately from the implementation
so that the evaluated instructions can be inspected without embedding several pages
of interface text in this appendix.

\subsection{Candidate Literature and Reference Resources}
\label{app:candidate_resources}
Codex and the two Deep Research systems use their native web search and source
selection. AutoSurvey and SurveyForge retain their released databases because those
resources are integral to their methods. For systems without a reproducible
retrieval source, we freeze the same 300 candidates selected for \method{} and
preserve the paper representation expected by the implementation. Naive RAG receives
only the 300 corresponding metadata records. This design avoids attributing
variation from unstable online acquisition to the downstream generation method.
\method{} itself permits a retrieval budget of up to 1,000 papers; the taxonomy and
routing experiments use the top 300 candidates.

\subsection{Baseline Configurations}
We use the official public implementation of every reproducible baseline and retain
its default control flow. Minimal adapters change the model endpoint or serialize
the fixed candidate input; they do not add \method{} planning, routing, review, or
visual modules. When a system outputs Markdown only, we render that output to PDF
without rewriting its content so that all methods can be evaluated from manuscript
pages. Codex was run on July 7, 2026 with GPT-5.5 Thinking. GPT Deep Research and
Gemini Deep Research were run from July 6 to July 8, 2026 with GPT-5.5 Thinking and
Gemini 3.1 Pro Deep Research, respectively.

\begin{table*}[t]
\centering
\small
\setlength{\tabcolsep}{3.0pt}
\begin{tabular}{llll}
\toprule
System & Generation model & Literature resource & Preserved system behavior \\
\midrule
Codex & GPT-5.5 Thinking & native search & search, selection, LaTeX, compilation \\
GPT DR & GPT-5.5 Thinking & native search & native Deep Research workflow \\
Gemini DR & Gemini 3.1 Pro DR & native search & native Deep Research workflow \\
Naive RAG & Qwen3.5-397B & fixed 300 metadata records & direct long-form generation \\
AutoSurvey & Qwen3.5-397B & released 530K corpus & official retrieval and survey pipeline \\
SurveyForge & Qwen3.5-397B & released paper and survey corpora & official outline and retrieval pipeline \\
LiRA & Qwen3.5-397B & fixed 300 candidates & official multi-role workflow \\
InteractiveSurvey & Qwen3.5-397B & fixed 300 candidates & official interactive RAG core \\
SurveyX & Qwen3.5-397B & fixed 300 candidates & official offline pipeline \\
\bottomrule
\end{tabular}
\caption{Baseline configurations. Qwen3.5-397B denotes Qwen3.5-397B-A17B-FP8.}
\label{tab:baseline_configs}
\end{table*}

\subsection{Human Reference Surveys}
The Human row contains one public academic survey matched to each \benchmark{}
topic. Candidates were collected from arXiv, journals, and conferences, with
preference given to work published from 2020 through June 15, 2026. We first
identified a primary and a backup survey for each topic and fixed the final choice
before scoring. Selection considered topical correspondence, survey completeness,
and compatibility with the evaluation inputs. Eligible surveys use numbered
citations, are generally no longer than 100 pages, and contain at least five
references from 2020--2026 that can be mapped to arXiv identifiers.

Matching considers the title, abstract, scope, and principal sections rather than
title keywords alone. When no survey exactly matches a benchmark title, we select
the candidate with the closest research object, technical scope, and central
questions. Human surveys are evaluated by the same 16 criteria and obtain a Total
Avg. of 4.34, equal to \method{} at the displayed precision. The Human row is a
quality reference, not a generated system, and is excluded from best and second-best
marking. Equality under \evalname{} does not imply that generated and human surveys
are interchangeable or require the same amount of expert revision.

{\footnotesize
\setlength{\tabcolsep}{3.0pt}
\begin{longtable}{@{}p{0.05\textwidth}p{0.52\textwidth}p{0.06\textwidth}p{0.27\textwidth}@{}}
\caption{Human-written surveys matched to the 30 \benchmark{} topics. The year and venue refer to the published version when available; otherwise, the initial arXiv year is reported.}
\label{tab:human_reference_surveys}\\
\toprule
ID & Matched human survey & Year & Venue/source \\
\midrule
\endfirsthead
\multicolumn{4}{l}{\textit{Table~\thetable{} continued.}}\\
\toprule
ID & Matched human survey & Year & Venue/source \\
\midrule
\endhead
\bottomrule
\endfoot
001 & Tool Learning with Large Language Models: A Survey \cite{human001_qu2025} & 2025 & Frontiers of Computer Science \\
002 & A Survey on the Memory Mechanism of Large Language Model-based Agents \cite{human002_zhang2025} & 2025 & ACM TOIS \\
003 & Retrieval-Augmented Generation for Large Language Models: A Survey \cite{human003_gao2023} & 2023 & arXiv \\
004 & Large Language Models for Planning: A Comprehensive and Systematic Survey \cite{human004_cao2025} & 2025 & arXiv \\
005 & The Landscape of Prompt Injection Threats in LLM Agents: From Taxonomy to Analysis \cite{human005_wang2026} & 2026 & arXiv \\
006 & A Systematic Literature Review on Large Language Models for Automated Program Repair \cite{human006_zhang2024} & 2024 & arXiv \\
007 & A Survey of Multimodal Retrieval-Augmented Generation \cite{human007_mei2025} & 2025 & arXiv \\
008 & Pure Vision Language Action (VLA) Models: A Comprehensive Survey \cite{human008_zhang2025} & 2025 & arXiv \\
009 & Text-to-image Diffusion Models in Generative AI: A Survey \cite{human009_zhang2023} & 2023 & arXiv \\
010 & Dynamic Scene Representation in the Era of Neural Rendering: From NeRFs to 3DGSs \cite{human010_han2026} & 2026 & Frontiers of Computer Science \\
011 & Multi-modal Sensor Fusion for Auto Driving Perception: A Survey \cite{human011_huang2022} & 2022 & arXiv \\
012 & Continual Learning of Large Language Models: A Comprehensive Survey \cite{human012_shi2025} & 2025 & ACM Computing Surveys \\
013 & A Survey on Offline Reinforcement Learning: Taxonomy, Review, and Open Problems \cite{human013_prudencio2024} & 2024 & IEEE TNNLS \\
014 & Graph Foundation Models: A Comprehensive Survey \cite{human014_wang2025} & 2025 & arXiv \\
015 & A Survey on Efficient Inference for Large Language Models \cite{human015_zhou2024} & 2024 & arXiv \\
016 & A Comprehensive Survey on Vector Database: Storage and Retrieval Technique, Challenge \cite{human016_ma2023} & 2023 & arXiv \\
017 & A Comprehensive Survey of Privacy-preserving Federated Learning: A Taxonomy, Review, and Future Directions \cite{human017_yin2021} & 2021 & ACM Computing Surveys \\
018 & LLMs in Software Security: A Survey of Vulnerability Detection Techniques and Insights \cite{human018_sheng2025} & 2025 & arXiv \\
019 & Transforming Science with Large Language Models: A Survey on AI-assisted Scientific Discovery, Experimentation, Content Generation, and Evaluation \cite{human019_eger2025} & 2025 & arXiv \\
020 & Toward Causal Representation Learning \cite{human020_scholkopf2021} & 2021 & Proceedings of the IEEE \\
021 & A Survey on Large Language Models for Recommendation \cite{human021_wu2024} & 2024 & World Wide Web \\
022 & Generative Modeling in Protein Design: Neural Representations, Conditional Generation, and Evaluation Standards \cite{human022_wanasekara2026} & 2026 & arXiv \\
023 & Single-cell Foundation Models: Bringing Artificial Intelligence into Cell Biology \cite{human023_baek2025} & 2025 & Experimental \& Molecular Medicine \\
024 & Survey on Deep Learning in Multimodal Medical Imaging for Cancer Detection \cite{human024_tian2023} & 2023 & Neural Computing and Applications \\
025 & Machine Learning Pipelines for the Design of Solid-State Electrolytes \cite{human025_jain2026} & 2026 & Materials Horizons \\
026 & Unlocking the Potential: Machine Learning Applications in Electrocatalyst Design for Electrochemical Hydrogen Energy Transformation \cite{human026_ding2024} & 2024 & Chemical Society Reviews \\
027 & Foundation Models for Weather and Climate Data Understanding: A Comprehensive Survey \cite{human027_chen2023} & 2023 & arXiv \\
028 & Foundation Models for Remote Sensing and Earth Observation: A Survey \cite{human028_xiao2025} & 2025 & IEEE Geoscience and Remote Sensing Magazine \\
029 & A Comprehensive Survey on Enterprise Financial Risk Analysis from Big Data and LLMs Perspective \cite{human029_du2022} & 2022 & arXiv \\
030 & A Survey on Uncertainty Quantification Methods for Deep Learning \cite{human030_he2023} & 2023 & arXiv; accepted by ACM Computing Surveys \\
\end{longtable}
}

\subsection{Timeout and Completion Protocol}
We impose a budget of 12 elapsed hours per topic and method, measured from process
launch. A run is completed only if it produces a nonempty, readable final PDF.
Retrieved papers, outlines, paper mappings, and partial drafts do not count as a
completed survey. A run without a valid PDF within the budget is recorded as a
timeout rather than assigned a quality score. Aggregate completion, time, page, and
reference statistics appear in \cref{app:completion_efficiency}.

\noindent\textbf{SurveyX record.}
We evaluated the public SurveyX offline pipeline under the fixed candidate setting
on five survey topics, none of which produced a final \texttt{survey.pdf} within the
12-hour budget. For the representative topic ``Tool Learning and Function Calling
for LLM Agents,'' eight attempts were archived. In the most complete run, outline
generation required 5,974.16 seconds. Paper mounting retained 296 records, with 280
nonempty mappings, 16 valid empty mappings, and no parsing failures. The content log
was last updated about 9 hours and 28 minutes after launch, after 5.56M input and
0.649M output tokens, but no raw main body, refined body, \texttt{survey.tex}, or
final PDF was written. We therefore classify all five SurveyX runs as timed out and
do not interpret these timeouts as manuscript quality failures.

\section{\evalname{} Rubrics and Evaluation Procedure}
\label{app:das_eval}
This section reproduces the complete scoring rubrics, evaluation inputs, judge
configurations, output validation rules, and score aggregation procedure cited as
Section E in the main paper. The executable judge prompts, payload builders, parsers,
and evaluation configuration files are provided in the evaluation directory of the
Code and Data Supplement. The descriptions below and these versioned artifacts
jointly supply the cited evaluation material.

\subsection{Metric Definitions and Scoring Anchors}
\evalname{} separates literature grounding, global survey organization, local
discourse, and visible manuscript reliability. Each submetric is scored with an
integer from 1 to 5. A score of 1 indicates absence or unusable quality, 2 a serious
defect, 3 competent but material limitations, 4 strong quality with localized
weaknesses, and 5 quality approaching the standard expected for submission in that
dimension. Judges must identify positive evidence and the strongest limitation
before assigning a score; manuscript length, citation count, and visual count do not
raise a score by themselves.

\begin{table*}[t]
\centering
\small
\setlength{\tabcolsep}{3.2pt}
\begin{tabular}{llp{10.7cm}}
\toprule
Group & Submetric & Operational question \\
\midrule
BSC & Citation Support & Do single citations and complete citation groups support their associated claims? \\
BSC & Attribution & Are tasks, methods, data, results, and limitations attributed to the correct papers? \\
BSC & Multi-Reference Synthesis & Do groups support explicit comparisons, families, trends, boundaries, or trade-offs? \\
BSC & Citation Balance & Are citations distributed without excessive concentration, repeated groups, or mechanical stacks? \\
\midrule
TSQ & Research Coverage & Does the manuscript substantively cover the major directions and research questions? \\
TSQ & Boundary Clarity & Are taxonomy branches coherent, distinct, and appropriately scoped? \\
TSQ & Global Organization & Do sections form a functional sequence rather than an unrelated topic list? \\
TSQ & Research Insights & Are comparisons, limitations, gaps, and prospects derived from the reviewed literature? \\
\midrule
HDQ & Goal Alignment & Do paragraphs and subsections serve their stated parent objectives? \\
HDQ & Argument Progression & Does local discussion advance through mechanisms, evidence, comparison, and consequence? \\
HDQ & Technical Specificity & Are claims concrete about methods, conditions, data, or evaluation rather than generic? \\
HDQ & Local Synthesis & Does prose integrate papers into bounded claims rather than enumerate them? \\
\midrule
MAR & Reference Integrity & Are visible citations and reference entries readable, consistent, complete, and traceable? \\
MAR & Visual Integration & Are meaningful figures and tables legible, captioned, numbered, and discussed in nearby text? \\
MAR & Layout Quality & Are typography, hierarchy, spacing, columns, equations, and page geometry stable? \\
MAR & Component Completeness & Are title, abstract, introduction, body, conclusion or outlook, and references complete? \\
\bottomrule
\end{tabular}
\caption{Operational definitions of the 16 \evalname{} submetrics.}
\label{tab:das_eval_metrics}
\end{table*}

\noindent\textbf{Complete criterion-level scoring rubrics.}
The complete anchors used by both judges are reported in
\cref{tab:bsc_rubric,tab:tsq_rubric,tab:hdq_rubric,tab:mar_rubric}. A score is
assigned only when the corresponding scoring description is satisfied.

{\footnotesize
\setlength{\tabcolsep}{4pt}
\begin{longtable}{@{}p{0.21\textwidth}cp{0.67\textwidth}@{}}
\caption{Complete scoring rubric for Balanced Scholarly Citation Quality (BSC).}
\label{tab:bsc_rubric}\\
\toprule
Submetric & Score & Scoring anchor \\
\midrule
\endfirsthead
\multicolumn{3}{l}{\textit{Table~\thetable{} continued.}}\\
\toprule
Submetric & Score & Scoring anchor \\
\midrule
\endhead
\bottomrule
\endfoot
Citation Support & 5 & Nearly all material claims are directly supported by the cited paper or complete citation group, with no consequential unsupported claim. \\
& 4 & Most material claims are directly supported; only a few localized claims have partial, indirect, or incomplete support. \\
& 3 & Core claims usually have relevant citations, but partial support, indirect support, or missing support recurs in the manuscript. \\
& 2 & Many central claims are weakly supported, and citation groups repeatedly contain insufficient or unrelated evidence. \\
& 1 & Citations are absent or unusable, or central claims are pervasively unsupported or contradicted by their cited evidence. \\
\midrule
Attribution & 5 & Paper-specific tasks, methods, data, results, contributions, and limitations are consistently attributed to the correct sources. \\
& 4 & Attribution is accurate overall, with only isolated imprecision that does not alter the main scholarly interpretation. \\
& 3 & The relevant papers are generally identified, but conflation, overstatement, or imprecise attribution appears repeatedly. \\
& 2 & Misattribution is frequent, or the manuscript repeatedly assigns unsupported details or conclusions to cited papers. \\
& 1 & Paper-specific attribution is fundamentally unreliable, fabricated, or untraceable. \\
\midrule
Multi-Reference Synthesis & 5 & Citation groups consistently support explicit comparisons, method families, trends, boundaries, common limitations, or trade-offs across papers. \\
& 4 & Strong synthesis appears across the major sections, with only a few missed opportunities or weakly developed comparisons. \\
& 3 & Meaningful synthesis is present, but it is mixed with paper-by-paper description and some citation groups lack a clear joint claim. \\
& 2 & The manuscript is dominated by isolated paper summaries or mechanical citation stacks with little explanation of cross-paper relations. \\
& 1 & No meaningful multi-reference synthesis is present. \\
\midrule
Citation Balance & 5 & Citations are distributed appropriately across claims, sections, and research directions, with minimal repetition or concentration. \\
& 4 & Coverage is generally balanced, with limited local concentration, repeated groups, or underrepresented directions. \\
& 3 & Several directions are represented, but noticeable concentration, repeated citation groups, or uneven section support remains. \\
& 2 & A small paper set is repeatedly reused, citation stacking is common, or major directions receive little scholarly support. \\
& 1 & Citations are largely absent, unusable, or overwhelmingly concentrated in a way that prevents balanced literature support. \\
\end{longtable}
}

{\footnotesize
\setlength{\tabcolsep}{4pt}
\begin{longtable}{@{}p{0.21\textwidth}cp{0.67\textwidth}@{}}
\caption{Complete scoring rubric for Taxonomic Synthesis Quality (TSQ).}
\label{tab:tsq_rubric}\\
\toprule
Submetric & Score & Scoring anchor \\
\midrule
\endfirsthead
\multicolumn{3}{l}{\textit{Table~\thetable{} continued.}}\\
\toprule
Submetric & Score & Scoring anchor \\
\midrule
\endhead
\bottomrule
\endfoot
Research Coverage & 5 & The survey substantively covers the major directions, technical branches, research questions, and representative developments of the topic. \\
& 4 & Most major directions and questions are covered, with only limited omissions or locally uneven depth. \\
& 3 & The central area is recognizable and several important directions are covered, but material gaps or strongly uneven treatment remain. \\
& 2 & Coverage is narrow or superficial, and multiple major directions or questions are absent. \\
& 1 & The manuscript does not establish a substantive account of the research space. \\
\midrule
Boundary Clarity & 5 & Taxonomy branches are coherent, appropriately scoped, and clearly distinguished, with well controlled overlap. \\
& 4 & Boundaries are clear overall, with only minor overlap, ambiguity, or uneven granularity. \\
& 3 & The taxonomy is usable, but recurring overlap, inconsistent scope, or misplaced material weakens several boundaries. \\
& 2 & Boundaries are frequently blurred or contradictory, causing substantial duplication or confusion. \\
& 1 & No coherent or defensible taxonomy is present. \\
\midrule
Global Organization & 5 & Sections have clear functions and form a coherent progression that supports understanding of the complete research area. \\
& 4 & The global sequence is coherent, with only a few weak transitions, redundant sections, or locally misplaced discussions. \\
& 3 & A recognizable organization exists, but some sections are list-like, weakly connected, redundant, or functionally unclear. \\
& 2 & The manuscript is fragmented, repeatedly redundant, or organized as loosely connected topics. \\
& 1 & The global organization is absent or prevents the manuscript from functioning as a survey. \\
\midrule
Research Insights & 5 & Evidence-based comparisons, trends, limitations, gaps, and prospects are developed throughout the survey and follow from the reviewed literature. \\
& 4 & Multiple sound insights are present across major sections, although their depth or distribution is somewhat uneven. \\
& 3 & Some useful analysis is present, but descriptive summary remains dominant and several insights are generic or weakly developed. \\
& 2 & Analysis is rare, generic, speculative, or insufficiently connected to the reviewed literature. \\
& 1 & The manuscript provides no meaningful research insight or presents conclusions that conflict with its literature discussion. \\
\end{longtable}
}

{\footnotesize
\setlength{\tabcolsep}{4pt}
\begin{longtable}{@{}p{0.21\textwidth}cp{0.67\textwidth}@{}}
\caption{Complete scoring rubric for Hierarchical Discourse Quality (HDQ).}
\label{tab:hdq_rubric}\\
\toprule
Submetric & Score & Scoring anchor \\
\midrule
\endfirsthead
\multicolumn{3}{l}{\textit{Table~\thetable{} continued.}}\\
\toprule
Submetric & Score & Scoring anchor \\
\midrule
\endhead
\bottomrule
\endfoot
Goal Alignment & 5 & Subsections and paragraphs consistently serve their stated parent objectives, with no material drift or misplaced discussion. \\
& 4 & Alignment is strong overall, with only isolated generic, tangential, or weakly placed passages. \\
& 3 & Most local units relate to their parent goals, but generic discussion, repetition, or partial drift recurs. \\
& 2 & Many paragraphs or subsections are weakly connected to their stated objectives or duplicate material assigned elsewhere. \\
& 1 & Local content is largely unrelated to the stated section goals or lacks a discernible objective. \\
\midrule
Argument Progression & 5 & Local discourse advances coherently through mechanisms, evidence, comparison, interpretation, and consequence where appropriate. \\
& 4 & Most paragraphs and subsection sequences progress logically, with only a few abrupt transitions or weak argumentative links. \\
& 3 & Individual paragraphs are generally coherent, but progression across paragraphs is uneven, repetitive, or mainly additive. \\
& 2 & Claims are loosely juxtaposed, transitions are frequently missing, or the discussion repeatedly resets without development. \\
& 1 & Local discourse is incoherent or does not form an argument. \\
\midrule
Technical Specificity & 5 & Claims consistently identify concrete mechanisms, assumptions, conditions, data, evaluation settings, results, or limitations. \\
& 4 & Technical discussion is mostly concrete, with only localized generic claims or missing conditions. \\
& 3 & Specific and technically useful material is present, but generic statements and omitted conditions recur. \\
& 2 & Most claims remain high level, vague, or detached from technical details needed for interpretation. \\
& 1 & The discussion is almost entirely generic, technically unusable, or materially incorrect. \\
\midrule
Local Synthesis & 5 & Paragraphs consistently integrate multiple papers into bounded claims, comparisons, shared mechanisms, conditions, or limitations. \\
& 4 & Local synthesis is strong in most analytical passages, with limited paper enumeration or weakly integrated evidence. \\
& 3 & Synthesis and paper-by-paper description coexist, and several paragraphs do not fully connect their cited works. \\
& 2 & Local writing is dominated by sequential paper summaries with little comparison or shared interpretation. \\
& 1 & Papers are merely listed, or no meaningful local synthesis is present. \\
\end{longtable}
}

{\footnotesize
\setlength{\tabcolsep}{4pt}
\begin{longtable}{@{}p{0.21\textwidth}cp{0.67\textwidth}@{}}
\caption{Complete scoring rubric for Manuscript Assembly Reliability (MAR).}
\label{tab:mar_rubric}\\
\toprule
Submetric & Score & Scoring anchor \\
\midrule
\endfirsthead
\multicolumn{3}{l}{\textit{Table~\thetable{} continued.}}\\
\toprule
Submetric & Score & Scoring anchor \\
\midrule
\endhead
\bottomrule
\endfoot
Reference Integrity & 5 & In-text citations and reference entries are readable, mutually consistent, complete, and traceable throughout the manuscript. \\
& 4 & Reference presentation is reliable overall, with only isolated minor inconsistencies or incomplete entries. \\
& 3 & The reference system remains usable, but several mismatches, unresolved items, malformed entries, or readability problems are visible. \\
& 2 & Invalid citations, missing entries, inconsistent numbering, or unreadable references recur and materially hinder tracing. \\
& 1 & The citation and reference system is absent, broken, or unusable. \\
\midrule
Visual Integration & 5 & Meaningful figures and tables are legible, correctly captioned and numbered, and explicitly integrated into nearby analytical discussion. \\
& 4 & Visual support is useful and well integrated overall, with only minor issues in legibility, placement, captioning, or textual discussion. \\
& 3 & Some visuals support the manuscript, but others are generic, weakly discussed, poorly placed, or only partly legible. \\
& 2 & Useful visual support is largely absent, or multiple figures and tables are unreadable, unreferenced, misleading, or disconnected from the text. \\
& 1 & No usable visual support is present, or visual artifacts substantially damage the manuscript. \\
\midrule
Layout Quality & 5 & Typography, hierarchy, spacing, columns, equations, tables, and page geometry remain stable and professional throughout. \\
& 4 & Layout is strong overall, with only isolated minor defects that do not impede reading. \\
& 3 & The manuscript is readable, but recurring minor defects or an isolated major defect reduces presentation quality. \\
& 2 & Frequent overflow, unstable spacing, broken columns, misplaced elements, or inconsistent typography materially impedes reading. \\
& 1 & Layout failures make substantial portions of the manuscript unreadable or unusable. \\
\midrule
Component Completeness & 5 & All major survey components are present, internally coherent, and sufficiently developed for a complete academic manuscript. \\
& 4 & All major components are present, but one is locally underdeveloped or only partially integrated. \\
& 3 & The manuscript is recognizable as a complete survey, but one major component is weak, incomplete, or substantially truncated. \\
& 2 & Multiple major components are absent, severely abbreviated, or disconnected from the manuscript. \\
& 1 & The output does not constitute a complete academic survey manuscript. \\
\end{longtable}
}

Scores 2 and 4 are not treated as generic intermediate values; they are assigned
only when their scoring descriptions are satisfied. The executable judge
prompts reproduce these anchors and are indexed by this section in the Code and Data
Supplement.

BSC treats supplied evidence cards as the only authority for paper content. Missing
evaluator metadata is marked unassessable and does not become a manuscript error.
TSQ and HDQ use method independent structural caps for outputs that are only thin
overviews, are dominated by references, or lack sufficient analytical depth. MAR
uses visual evidence at the page level and applies explicit caps for broken references,
unreadable or absent synthesis aids, recurring layout defects, and missing manuscript
components. These caps prevent a polished first page or large bibliography from
masking defects in the complete output.

\subsection{Evaluation Inputs and Citation Metadata}
Each final PDF is rendered at 160 dpi into one PNG per page and is also parsed into
Markdown. BSC receives JSONL records containing global citation statistics, a
deduplicated evidence card bank, cited claim contexts with complete citation groups,
and a balance summary computed from every parseable citation marker in the body.
Each visible citation number is mapped to an arXiv identifier through the archived
\texttt{ref.json}; mappings are checked against the final reference order. A missing
abstract or metadata field affects assessability, not the citation score by default.

TSQ and HDQ receive the complete manuscript PDF in API mode or its rendered page
images in local mode; the judge does not receive a separate deterministic audit
payload. MAR receives every rendered page for manuscripts of at most 50 pages. For
longer manuscripts, the evaluator keeps 20\% of the page budget from the front,
10\% from the back, and samples the remainder uniformly from the middle. The model
name is not included in the evaluation input. Diagnostics and rationales are
retained for audit, while only validated integer scores enter aggregation.

\subsection{LLM Judge Configuration}
The main judge is Qwen3.5-397B-A17B-FP8 with temperature 0.2 and a 600-second
request timeout. Kimi K2.6 is used for the independent cross-judge analysis with
temperature 0.6 and a 1,800-second timeout. Both are accessed through an
OpenAI-compatible API. Qwen requests permit three attempts with a $5a$ second delay
after failed attempt $a$; Kimi permits eight attempts with an $8a$ second delay.
The evaluation program supports resumption, and a failed request is never converted
to a numerical score.

\begin{lstlisting}[caption={Condensed overview of the judge contract.},label={lst:judge_prompt},captionpos=t,frame=tb,numbers=none,xleftmargin=0pt]
You are an expert reviewer evaluating an anonymous academic survey. Use only the supplied manuscript pages, parsed text, citation metadata, and evidence cards. Do not use external knowledge and do not infer the generating method.

Evaluate only the requested DAS-Eval dimensions. Inspect evidence across the complete input, including early, middle, late, visual, and reference regions where applicable. For every submetric, identify the strongest positive evidence and the strongest verified limitation. Do not reward document length, citation count, figure count, or visual ornament by itself.

Score each requested submetric with one integer:
5 = consistently excellent and close to submission quality;
4 = strong, with localized minor defects;
3 = competent but materially imperfect;
2 = weak, with serious recurring defects;
1 = absent, failed, or unusable.

Apply the metric definitions and score caps in the metric table above. For BSC, treat evidence cards as the only authority and mark missing evaluator evidence as unassessable. For TSQ and HDQ, judge research organization and discourse rather than PDF appearance. For MAR, judge only visible artifact quality, not scientific correctness.

Return only valid JSON. For each submetric return {"score": integer, "rationale": string}; include diagnostics and an overall assessment. Scores must be in [1,5], all required fields must be present, and each group total must equal the sum of its four submetrics.

Topic: {topic}
Requested dimensions: {dimensions}
Evaluation input:
{evaluation_payload}
\end{lstlisting}

No separate system message is used for \evalname{}. Each evaluator receives one
metric-specific user message that combines the corresponding scoring anchors,
evaluation input, diagnostic fields, hard caps, and JSON output schema. The runtime
uses separate BSC, TSQ/HDQ, and MAR requests so that citation evidence, discourse
assessment, and visual artifact criteria remain isolated. The complete archived
messages and parsing schemas are supplied in the Code and Data Supplement; the
static metric definitions used by those messages are reproduced in
\cref{tab:das_eval_metrics}.

\subsection{Score Aggregation and Missing Outputs}
Let $s_{m,t,c}\in\{1,\ldots,5\}$ be the score of method $m$ on topic $t$ and
submetric $c$. The reported submetric mean is
\begin{equation}
\bar{s}_{m,c}=\frac{1}{N_m}\sum_{t=1}^{N_m}s_{m,t,c},
\label{eq:submetric_average}
\end{equation}
where $N_m$ is the number of topics covered by the method. Most methods cover 30
topics; AutoSurvey and SurveyForge use the 21 CS topics supported by their released
corpora. For group $g$ and the overall score,
\begin{equation}
\begin{aligned}
\operatorname{GroupAvg}_{m,g}&=\frac{1}{4}\sum_{c\in g}\bar{s}_{m,c},\\
\operatorname{TotalAvg}_{m}&=\frac{1}{16}\sum_{c=1}^{16}\bar{s}_{m,c}.
\end{aligned}
\label{eq:score_aggregation}
\end{equation}

Aggregation uses unrounded values, and rounding is applied once for display. Best
and second-best generated systems are determined from unrounded means. Human
surveys are excluded from these markings. Exact ties share a rank; values that only
appear tied after display rounding retain their unrounded order. The parser removes
optional Markdown fences, first attempts direct JSON parsing, and otherwise extracts
the first balanced JSON object. Known aliases and recoverable numeric forms are
normalized, and documented neutral fallbacks are used only for specific recoverable
omissions. Responses without a valid structured score are retried. Group totals are
recomputed rather than trusted from the model. The final main run uses 15 concurrent
evaluation tasks and permits up to 15 task submissions; all 846 metric results are
present. A generation timeout is not assigned zero and is excluded according to
\cref{app:completion_efficiency}.

\noindent\textbf{Validation of aggregated outputs.}
The result builder verifies the expected topic set, metric keys, score range, and
group arithmetic before producing a table. Any malformed, incomplete, or timed out
judge request is resubmitted through the resumable evaluation queue rather than
silently removed or mean imputed.

\section{Additional Experimental Results}
\label{app:additional_results}

\subsection{Matched Comparison on the Shared CS Topics}
\label{app:matched_cs_results}
\begin{table*}[t]
\centering
\small
\renewcommand{\arraystretch}{1.0}
\setlength{\tabcolsep}{1.0pt}
\resizebox{\textwidth}{!}{%
\begin{tabular}{l ccccc ccccc ccccc ccccc c}
\toprule
\raisebox{-1.3ex}[0pt][0pt]{Method}
& \multicolumn{5}{c}{BSC$\uparrow$}
& \multicolumn{5}{c}{TSQ$\uparrow$}
& \multicolumn{5}{c}{HDQ$\uparrow$}
& \multicolumn{5}{c}{MAR$\uparrow$}
& \raisebox{-1.3ex}[0pt][0pt]{Total} \\
\cmidrule(lr){2-6}\cmidrule(lr){7-11}\cmidrule(lr){12-16}\cmidrule(lr){17-21}
& Sup. & Attr. & MSyn. & Bal. & Avg.
& Cov. & Bnd. & Org. & Ins. & Avg.
& Aln. & Prog. & Spec. & LSyn. & Avg.
& Ref. & Vis. & Lay. & Comp. & Avg. & \\
\midrule
Human & 3.86&3.90&3.67&4.14&3.89 &4.29&4.38&4.43&4.38&4.37 &4.43&4.14&4.38&4.14&4.27 &5.00&5.00&5.00&5.00&5.00 &4.38\\
\midrule
Codex &3.00&3.33&2.76&2.05&2.79 &3.00&3.00&3.00&2.95&2.99 &3.48&2.48&3.05&2.00&2.75 &\textbf{5.00}&1.81&\textbf{5.00}&\textbf{5.00}&4.20 &3.18\\
GPT DR &3.67&3.76&3.48&2.71&3.40 &3.48&3.52&3.57&3.48&3.51 &4.14&3.52&4.00&3.52&3.80 &\textbf{5.00}&1.57&\textbf{5.00}&\textbf{5.00}&4.14 &3.71\\
Gemini DR &2.95&2.86&2.81&\underline{3.62}&3.06 &3.81&3.86&3.86&3.76&3.82 &4.14&\underline{3.95}&4.14&3.95&4.05 &\underline{4.43}&4.81&\underline{4.86}&\textbf{5.00}&4.77 &3.93\\
Naive RAG &\textbf{4.00}&\underline{4.00}&3.19&3.48&3.67 &\underline{3.95}&\textbf{4.19}&\underline{4.10}&3.76&\underline{4.00} &\textbf{4.33}&3.81&\underline{4.48}&3.81&\underline{4.11} &\textbf{5.00}&1.29&\textbf{5.00}&\textbf{5.00}&4.07 &\underline{3.96}\\
AutoSurvey &\textbf{4.00}&\underline{4.00}&3.29&\textbf{3.95}&\underline{3.81} &\textbf{4.00}&3.86&3.10&\underline{4.00}&3.74 &4.10&3.14&4.24&3.29&3.69 &\textbf{5.00}&1.29&\textbf{5.00}&3.38&3.67 &3.73\\
SurveyForge &\textbf{4.00}&3.90&\underline{3.81}&3.19&3.73 &\textbf{4.00}&\underline{4.00}&3.24&\underline{4.00}&3.81 &4.10&3.29&4.19&3.38&3.74 &\textbf{5.00}&1.29&\textbf{5.00}&\underline{4.05}&3.83 &3.78\\
LiRA &\underline{3.95}&3.95&3.62&3.00&3.63 &\textbf{4.00}&3.29&3.52&\underline{4.00}&3.70 &4.00&3.86&4.38&3.95&4.05 &3.00&1.33&4.81&3.00&3.04 &3.60\\
InteractiveSurvey &3.10&3.10&2.05&3.05&2.82 &\textbf{4.00}&3.76&3.86&\underline{4.00}&3.90 &3.81&\underline{3.95}&4.00&\underline{4.00}&3.94 &\underline{4.43}&\underline{4.86}&\underline{4.86}&\textbf{5.00}&\underline{4.79} &3.86\\
\midrule
\textbf{\method{}} &\textbf{4.00}&\textbf{4.05}&\textbf{4.00}&3.43&\textbf{3.87} &\textbf{4.00}&\underline{4.00}&\textbf{4.29}&\textbf{4.43}&\textbf{4.18} &\underline{4.24}&\textbf{4.10}&\textbf{4.52}&\textbf{4.14}&\textbf{4.25} &\textbf{5.00}&\textbf{5.00}&\textbf{5.00}&\textbf{5.00}&\textbf{5.00} &\textbf{4.32}\\
\bottomrule
\end{tabular}%
}
\caption{Matched comparison on the 21 CS topics shared by all completed systems. Best and second-best system scores are bold and underlined, while Human is reported for reference only.}
\label{tab:appendix_matched_cs_results}
\end{table*}

The comparison on shared topics preserves the ordering of generated systems by Total
Avg. reported in the main analysis: \method{} ranks first at 4.32, followed by
Naive RAG at 3.96 and Gemini Deep Research at 3.93. The result indicates that the
main ranking is not produced by the additional nine interdisciplinary topics.

\subsection{Component Ablation Details}
All component variants reuse the same 30 topics, candidate budget, generation model,
decoding settings, prompts outside the removed component, and output requirements.
\textit{w/o Metadata} replaces the structured record with title and abstract at all
online stages. \textit{w/o Taxonomy} generates the taxonomy from the topic without
candidate metadata. \textit{w/o Routing} replaces global paper assignment with
independent top-$k$ retrieval for each section. \textit{w/o Hierarchical Drafting}
generates each subsection in one pass while retaining its taxonomy objective and
candidate papers. \textit{w/o Deterministic Validation} removes paragraph checks and
rule repair. \textit{w/o Semantic Review} returns the initial complete subsection
without semantic critique or repair.

The full system and all variants except \textit{w/o Deterministic Validation}
compile 30/30 manuscripts. The variant without deterministic validation compiles
27/30 manuscripts. This deterministic outcome complements the
saturated MAR scores reported in Table~3 of the main paper.

\subsection{Repair Policy Analysis}
The repair policies start from identical subsection checkpoints saved immediately
before semantic review. Each checkpoint contains the initial draft, paragraph plan,
claim and citation plans, optional evidence, and paragraph states. The policies use
the same model, prompt, decoding, review budget, and deterministic checker. They
differ only in the action allowed after a non-PASS decision; the no-review variant
returns the shared initial draft.

\cref{tab:repair_actions} reports the average number and relative frequency of final
reviewer decisions per survey. Direct editing is the most frequent repair action,
followed by section and paragraph replanning. The observed policy therefore uses all
three repair scopes rather than relying on a single action.

\begin{table}[!t]
\centering
\small
\renewcommand{\arraystretch}{1.15}
\setlength{\tabcolsep}{3.5pt}
\begin{tabularx}{\textwidth}{@{}l*{4}{>{\centering\arraybackslash}X}@{}}
\toprule
& PASS & \makecell{REPLAN\_\\SECTION} & \makecell{REPLAN\_\\PARAGRAPH} & \makecell{DIRECT\_\\EDIT} \\
\midrule
Average selections per survey & 40.40 & 12.40 & 2.60 & 35.00 \\
Share of decisions & 44.69\% & 13.72\% & 2.88\% & 38.72\% \\
\bottomrule
\end{tabularx}
\caption{Average distribution of final reviewer decisions. Percentages are independently rounded.}
\label{tab:repair_actions}
\end{table}

\subsection{Completion, Efficiency, and Output Statistics}
\label{app:completion_efficiency}
\cref{tab:completion_stats} reports the final completion and output statistics.
Runtime is measured from launch to the final artifact for a successful run. For
timestamped batch queues, gaps exceeding two hours are treated as external
interruptions and excluded from continuous segments. Codex ran serially; its
normalized throughput is the elapsed interval from the second to the twenty-eighth
completed topic divided by 26 transitions. \method{} time is measured from each run
start to its original medium-length PDF under five concurrent runs. GPT and Gemini
Deep Research lack auditable timestamps for the reported generation run, so their
time entries remain unspecified rather than using vendor estimates.

Numbered references count distinct entries in the final reference section, and the
ratio divides that count by the requested target of 300. For continuously numbered
lists, we use the final index; for sparse lists, we count distinct reference IDs.
This measure remains comparable when PDF conversion removes brackets from in-text
citations. A ratio above 100\% indicates that a system includes more than 300
numbered references rather than being truncated at the requested target. Completion
is computed over the topics supported by each method: 21 for AutoSurvey and
SurveyForge, 30 for the other completed systems, and five attempted topics for
SurveyX.

\begin{table}[!t]
\centering
\small
\setlength{\tabcolsep}{4.0pt}
\begin{tabular}{lrrrrrr}
\toprule
System & Time (h) & Timeouts & Completion & Pages & Numbered refs. & Refs./300 \\
\midrule
Codex & 0.83 & 0 & 100\% & 20.53 & 286.13 & 95.38\% \\
GPT DR & -- & 0 & 100\% & 15.47 & 132.70 & 44.23\% \\
Gemini DR & -- & 0 & 100\% & 19.23 & 99.17 & 33.06\% \\
Naive RAG & 0.20 & 0 & 100\% & 13.70 & 88.90 & 29.63\% \\
AutoSurvey & 0.94 & 0 & 100\% & 112.24 & 544.05 & 181.35\% \\
SurveyForge & 0.49 & 0 & 100\% & 30.62 & 157.71 & 52.57\% \\
LiRA & 3.58 & 0 & 100\% & 27.40 & 100.67 & 33.56\% \\
InteractiveSurvey & 5.25 & 0 & 100\% & 22.80 & 19.23 & 6.41\% \\
SurveyX & -- & 5 & 0\% & -- & -- & -- \\
\midrule
\textbf{\method{}} & 1.49 & 0 & 100\% & 47.20 & 217.83 & 72.61\% \\
\bottomrule
\end{tabular}
\caption{Completion, efficiency, and output statistics. Time is the mean number of elapsed hours per completed survey under the stated execution protocol. Dashes denote unavailable runtime measurements or output statistics that are inapplicable to timed-out runs.}
\label{tab:completion_stats}
\end{table}

All completed systems produce the full supported topic set after permitted reruns.
LiRA has one failed attempt that is rerun successfully and is not counted as a
timeout. SurveyX is the only system that fails to produce a complete PDF within the
12-hour budget for all five attempted surveys (0/5), so it is excluded from the
output quality comparison; a representative run record is detailed in
\cref{app:das_bench}. Runtime values should
be interpreted as observed system throughput rather than a pure model speed
comparison because closed and open systems execute in different environments.

\section{Human Evaluation and Cross-Judge Robustness}
\label{app:human_judge}

\subsection{Expert Recruitment and Assignment}
In the main paper, \emph{domain experts} refers to expert evaluators with research
experience in language agents, natural language processing, related areas of
artificial intelligence, and academic survey assessment. It does not imply that a
separate subject specialist was recruited for every interdisciplinary benchmark
topic. We recruited three such external evaluators: two senior doctoral researchers
and one research scientist. None is an author of this paper, a member of the
authors' laboratory, or a member of a collaborating team. None participated in the
development of \method{}, generation of the evaluated manuscripts, or automatic
evaluation. The evaluators assessed literature support, organization, analytical
writing, and manuscript completeness; they were not asked to independently verify
every technical claim in every benchmark domain.

All three evaluators assessed every available topic. The 21 CS sets contain
\method{}, Naive RAG, and AutoSurvey. The nine non-CS sets contain \method{} and
Naive RAG because AutoSurvey does not provide outputs beyond its supported CS
corpus. The resulting design contains $21\times3=63$ CS and $9\times2=18$ non-CS
topic--method outputs, yielding $21\times3+9=72$ within-topic pairwise
comparisons.

\subsection{Blind Ranking Protocol}
For each topic, method names and file identifiers associated with each method are removed from
the complete PDFs. CS outputs are presented as \texttt{A}, \texttt{B}, and
\texttt{C}; non-CS outputs use \texttt{A} and \texttt{B}. The label mapping varies
across topics and is stored separately. Topic order is shuffled. Evaluators do not
receive the mapping or automatic scores before submitting their decisions.

Each evaluator ranks manuscripts only within a topic. Ties are permitted when no
meaningful overall quality difference can be identified, but are not used merely
because multiple outputs are acceptable. Complete rankings are converted to
pairwise win, tie, or loss labels. The majority label is the judgment selected by at
least two experts. All 72 pairwise comparisons yield a valid majority and no cyclic
majority occurs in a three-system topic.

\begin{lstlisting}[caption={Instructions provided to the external evaluators.},label={lst:expert_instructions},captionpos=t,frame=tb,numbers=none,xleftmargin=0pt]
You will evaluate anonymous academic survey manuscripts. For each topic, read the complete PDFs and rank them by their overall quality as publication-oriented academic surveys.

Consider jointly:
1. whether important academic claims are adequately supported and attributed to relevant literature;
2. whether the taxonomy, section structure, and organization coherently cover the topic;
3. whether the manuscript analyzes and compares papers rather than merely listing prior work; and
4. whether the manuscript is complete, readable, and presented as an academic survey.

Provide one overall ranking per topic. Use A>B>C when A is preferred to B and B to C. Ties such as A=B>C are allowed only when there is no meaningful overall quality difference. Evaluate topics independently. Do not infer system identities, consult another evaluator, compare manuscripts across topics, or revise a decision using another evaluator's judgment.
\end{lstlisting}

\subsection{Complete Expert Results}
We denote \method{} by D, Naive RAG by R, AutoSurvey by A, and the evaluators by
E1--E3. \cref{tab:expert_rankings} reports all judgments.

\begin{table}[!t]
\centering
\small
\setlength{\tabcolsep}{5.0pt}
\begin{tabular}{ccccc@{\qquad}ccccc}
\toprule
Topic & E1 & E2 & E3 & Majority & Topic & E1 & E2 & E3 & Majority \\
\midrule
1&D=A$>$R&D$>$A$>$R&D$>$A$>$R&D$>$A$>$R &16&D$>$A$>$R&D$>$A$>$R&D$>$A$>$R&D$>$A$>$R\\
2&D$>$R$>$A&D$>$A$>$R&D$>$A$>$R&D$>$A$>$R &17&D$>$A$>$R&D$>$A$>$R&D$>$A$>$R&D$>$A$>$R\\
3&D$>$A$>$R&D$>$A$>$R&D$>$A$>$R&D$>$A$>$R &18&R$>$A$>$D&R$>$A$>$D&R=A$>$D&R$>$A$>$D\\
4&D$>$R$>$A&D$>$R$>$A&D$>$R$>$A&D$>$R$>$A &19&D$>$R=A&D$>$R$>$A&D$>$R$>$A&D$>$R$>$A\\
5&D$>$A$>$R&D$>$A$>$R&D$>$A$>$R&D$>$A$>$R &20&R$>$A$>$D&R$>$A$>$D&R=A$>$D&R$>$A$>$D\\
6&D$>$A$>$R&D$>$A$>$R&D$>$A$>$R&D$>$A$>$R &21&D$>$A$>$R&D$>$R=A&D$>$A$>$R&D$>$A$>$R\\
7&D$>$A$>$R&D$>$A$>$R&D$>$A$>$R&D$>$A$>$R &22&D$>$R&D$>$R&D$>$R&D$>$R\\
8&D$>$A$>$R&D$>$A$>$R&D$>$A$>$R&D$>$A$>$R &23&D$>$R&D$>$R&D$>$R&D$>$R\\
9&R$>$D$>$A&R$>$D$>$A&R$>$D$>$A&R$>$D$>$A &24&D$>$R&D$>$R&D$>$R&D$>$R\\
10&D$>$R$>$A&D$>$R$>$A&D$>$R$>$A&D$>$R$>$A &25&D$>$R&D$>$R&D$>$R&D$>$R\\
11&D$>$A$>$R&D$>$A$>$R&D$>$A$>$R&D$>$A$>$R &26&D$>$R&D$>$R&D$>$R&D$>$R\\
12&R$>$D$>$A&D$>$A$>$R&D$>$A$>$R&D$>$A$>$R &27&D$>$R&D$>$R&D$>$R&D$>$R\\
13&D$>$A$>$R&D$>$A$>$R&D$>$A$>$R&D$>$A$>$R &28&D$>$R&D$>$R&D$>$R&D$>$R\\
14&D$>$A$>$R&D$>$A$>$R&D$>$A$>$R&D$>$A$>$R &29&D$>$R&D$>$R&D$>$R&D$>$R\\
15&D$>$R=A&D$>$A$>$R&D$>$A$>$R&D$>$A$>$R &30&D$>$R&D$>$R&D$>$R&D$>$R\\
\bottomrule
\end{tabular}
\caption{Complete blind rankings from the three external evaluators. D, R, and A denote \method{}, Naive RAG, and AutoSurvey.}
\label{tab:expert_rankings}
\end{table}

\begin{table}[!t]
\centering
\footnotesize
\setlength{\tabcolsep}{3.6pt}
\begin{tabular}{lccc}
\toprule
Comparison & Topics & Majority W--T--L & Individual W--T--L \\
\midrule
D vs. R & 30 & 27--0--3 & 80--0--10 \\
D vs. A & 21 & 19--0--2 & 56--1--6 \\
A vs. R & 21 & 15--0--6 & 41--5--17 \\
\bottomrule
\end{tabular}
\caption{Pairwise expert preferences. Win, tie, and loss are stated for the method on the left.}
\label{tab:expert_pairwise}
\end{table}

Under majority judgments, \method{} ranks first on 18 of 21 CS topics and is
preferred to Naive RAG on 27 of all 30 topics. All three experts give identical
pairwise labels for 63 of 72 comparisons, an agreement rate of 87.5\%. Fleiss'
$\kappa=0.726$, indicating substantial agreement beyond chance; the other nine
comparisons still have a two-expert majority. Seven of these nine disagreements
concern AutoSurvey versus Naive RAG. Pairwise Cohen's $\kappa$ is 0.700 for E1--E2,
0.650 for E1--E3, and 0.848 for E2--E3.

\subsection{Cross-Judge Agreement}
\begin{table}[t]
\centering
\small
\setlength{\tabcolsep}{2.2pt}
\resizebox{\textwidth}{!}{%
\begin{tabular}{ll cccc cccc cccc cccc c}
\toprule
Judge & Method
& \multicolumn{4}{c}{BSC}
& \multicolumn{4}{c}{TSQ}
& \multicolumn{4}{c}{HDQ}
& \multicolumn{4}{c}{MAR}
& Total \\
\cmidrule(lr){3-6}\cmidrule(lr){7-10}\cmidrule(lr){11-14}\cmidrule(lr){15-18}
& & Sup. & Attr. & MSyn. & Bal.
& Cov. & Bnd. & Org. & Ins.
& Aln. & Prog. & Spec. & LSyn.
& Ref. & Vis. & Lay. & Comp. & \\
\midrule
Qwen3.5 & Naive RAG
&4.00&4.00&3.19&3.48 &3.95&4.19&4.10&3.76 &4.33&3.81&4.48&3.81 &5.00&1.29&5.00&5.00 &3.96\\
Qwen3.5 & AutoSurvey
&4.00&4.00&3.29&3.95 &4.00&3.86&3.10&4.00 &4.10&3.14&4.24&3.29 &5.00&1.29&5.00&3.38 &3.73\\
Qwen3.5 & \method{}
&4.00&4.05&4.00&3.43 &4.00&4.00&4.29&4.43 &4.24&4.10&4.52&4.14 &5.00&5.00&5.00&5.00 &4.32\\
\midrule
Kimi K2.6 & Naive RAG
&4.00&4.00&3.05&3.81 &3.00&3.00&3.00&2.95 &3.00&2.38&3.29&2.38 &3.86&2.00&3.95&3.95 &3.23\\
Kimi K2.6 & AutoSurvey
&4.00&4.00&3.00&3.57 &3.81&3.00&3.24&3.29 &3.48&3.10&3.81&3.10 &2.10&1.52&2.29&2.48 &3.11\\
Kimi K2.6 & \method{}
&3.95&3.90&3.71&3.10 &4.14&4.14&4.05&4.14 &4.14&4.00&4.14&4.00 &4.14&4.05&4.05&4.95 &4.04\\
\bottomrule
\end{tabular}%
}
\caption{Submetric averages under the main and independent judges on the shared 21-topic CS subset.}
\label{tab:cross_judge_results}
\end{table}

The main paper reports an analysis of all available outputs in which \method{} and Naive
RAG are averaged over 30 topics and AutoSurvey over its 21 supported CS topics.
Under that coverage, the two judges have Spearman $\rho=0.507$ and mean absolute
error 0.630 across the 48 method--submetric averages. To remove variation in topic
coverage, \cref{tab:cross_judge_results} recomputes all averages on the shared
21-topic CS subset, yielding $\rho=0.501$ and mean absolute error 0.617. Both
analyses preserve the ranking \method{} $>$ Naive RAG $>$ AutoSurvey. The judges
also agree on 48 of 63 topic-level comparisons on the CS subset. On the nine non-CS
topics, both rank \method{} above Naive RAG in aggregate, although topic-level
agreement is 5 of 9. These results support the main ordering but also show that
fine-grained scores remain sensitive to the judge model.

\section{Reproducibility, Release, and Limitations}
\label{app:reproducibility}

\subsection{Artifact and Release Plan}
We plan to release the \method{} source code, configuration templates, generation
and evaluation scripts, the \benchmark{} topic list, all \evalname{} rubrics and
judge prompts, the \dataset{} schema, distributable metadata records with canonical
arXiv identifiers, and anonymized generated outputs where redistribution is
permitted. The release will also contain the static website
described in \cref{app:supplementary_outputs}, result tables with unrounded values,
the citation mapping format, and scripts for rebuilding all reported aggregates.
Versioned configuration files will identify the benchmark snapshot and model names;
credentials, internal service endpoints, and copyrighted source PDFs will not be
distributed.

\subsection{Limitations}
The experimental literature snapshot covers arXiv papers from 2020 through June 2026 and
therefore underrepresents older work, non-arXiv venues, books, and domain databases.
Metadata extraction can omit or misstate technical details even when its structure
is valid; focused source access mitigates missing details but does not guarantee
that every extracted value is correct. The current system also depends on a large
generation model and substantial offline computation.

Closed research systems cannot be controlled as precisely as public implementations,
and their retrieval behavior may change over time. Only nine benchmark topics are
outside computer science. The backbone analysis covers five topics because of model
cost and should be interpreted as sensitivity evidence rather than a comprehensive
scaling study. Automatic scores vary across judge families, and the moderate
cross-judge correlation limits fine-grained conclusions. Finally, a
publication-oriented survey remains a draft: relevant subject specialists must
verify technical claims, literature coverage, attribution, and conclusions before
submission.

The metadata audit and cross-judge analysis should be understood as bounded
diagnostics rather than exhaustive guarantees. The former estimates source
alignment from a stratified sample, whereas the latter measures consistency between
two judge families under a fixed rubric. Future evaluation should broaden manual
source verification, include additional judge families, and cover more
interdisciplinary topics and generation backbones. Such extensions would better
characterize residual uncertainty without changing the controlled comparisons
reported in this study.

\end{document}